\documentclass{article}

\usepackage{microtype}
\usepackage{graphicx}
\usepackage{subcaption}
\usepackage{booktabs} 

\usepackage{amsfonts}       
\usepackage{nicefrac}       
\usepackage{microtype}      
\usepackage[dvipsnames]{xcolor}         
\usepackage{tikz}
\usetikzlibrary{calc}
\usetikzlibrary{arrows,scopes,decorations.pathreplacing,decorations.markings,shapes}
\usepackage{float}
\usepackage{amssymb}
\usepackage{wrapfig}
\usepackage{GLAGraphsvPP} 
\definecolor{classc}{RGB}{255,0,255}
\definecolor{workc}{RGB}{102,0,204}
\definecolor{trailc}{RGB}{0,204,0}
\definecolor{cityc}{RGB}{0,102,255}
\definecolor{calmingcolor}{RGB}{255, 145, 175} 
\usepackage{relsize}
\usepackage{moresize}
\usepackage{rotating}
\usepackage{dblfloatfix} 
\usepackage{epstopdf}

\usepackage{multirow}

\newcommand{\classcolor}[1]{\textcolor{classc}{#1}}
\newcommand{\workcolor}[1]{\textcolor{workc}{#1}}
\newcommand{\trailcolor}[1]{\textcolor{trailc}{#1}}
\newcommand{\citycolor}[1]{\textcolor{cityc}{#1}}

\newcommand{\finalSolve}[1]{\textcolor{black}{#1}}

\usepackage{xfrac}
\newcommand{\numsymbol}{\raisebox{-1.25pt}{\scalebox{1}{\#}}}

\newcommand{\M}[1]{\({#1}\)}

\newcommand{\ParagraphVertSpace}{\vspace{8pt}}

\newcommand{\TimesGap}{\kern3pt}
\newcommand{\MuShrinkGap}{\kern-1.5pt}

\newcommand{\Data}{\mathcal{D}}

\newcommand{\Evidence}{\mathcal{E}}
\newcommand{\EvidenceOne}{e}
\newcommand{\evidenceWord}{$\EvidenceOne\text{vidence}$}

\newcommand{\EvidenceWord}{$\Evidence\text{vidence}$}

\newcommand{\Hypo}{\mathcal{H}}
\newcommand{\HypoOne}{h}
\newcommand{\hypothesisWord}{$\HypoOne\text{ypothesis}$}
\newcommand{\hypothesesWord}{$\HypoOne\text{ypotheses}$}

\newcommand{\HypothesesWord}{$\Hypo\text{ypotheses}$}

\newcommand{\InfoIn}{\Evidence}
\newcommand{\InfoInOne}{\EvidenceOne}
\newcommand{\InfoInWord}{\evidenceWord}
\newcommand{\InfoInWordP}{\evidenceWord}

\newcommand{\InfoInWordPCap}{\EvidenceWord}

\newcommand{\InfoOut}{\Hypo}
\newcommand{\InfoOutOne}{\HypoOne}
\newcommand{\InfoOutWord}{\hypothesisWord}
\newcommand{\InfoOutWordP}{\hypothesesWord}

\newcommand{\InfoOutWordPCap}{\HypothesesWord}

\newcommand{\experimentalist}{\textsc{Expmnt}}
\newcommand{\participant}{\textsc{Ptcpnt}}
\newcommand{\ExpMapping}{\experimentalist}
\newcommand{\ParticipantMapping}{\participant}

\newcommand{\fractionObscure}{b}

\newcommand{\numShown}{s}

\newcommand{\numObs}{\numsymbol_{\text{obs}}}

\newcommand{\numRelations}{\numsymbol_{\text{all relations}}}

\newcommand{\PartialGraph}{P{\kern-2.0pt}G}
\newcommand{\PartialGraphSet}{\mathcal{\PartialGraph}}
\newcommand{\PartialGraphText}{$\PartialGraph$}

\newcommand{\avgLL}{\textsc{avgLL}}

\newcommand{\Graph}{G}
\newcommand{\GraphSet}{\mathcal{\Graph}}
\newcommand{\GraphText}{$\Graph$}

\newcommand{\Prior}{\pi}
\newcommand{\spacequote}{\vspace{6pt}}
\newcommand{\spacepageinstruction}{\vspace{8pt}}
\newcommand{\tableFontSocNav}[1]{#1}

\newcommand{\tableFontEntries}[1]{\text{#1}}

\newcommand{\tableFontEntriesNum}[1]{$#1$}

\newcommand{\tableFontStoryName}[1]{\textbf{\textsc{#1}}}

\newcommand{\tableFontTop}[1]{\textsc{#1}}
\newcommand{\tableFontTopBefAf}[1]{\text{#1}}
\newcommand{\tableFontStoryNameBefAft}[1]{\textbf{\textsc{#1}}}

\newcommand{\tablenumdatapropfont}[2]{{$#2$} out of {$#1$}}
\newcommand{\tablenumdataperc}[1]{\hspace{4pt}($#1 \%$)}

\usepackage{bm}
\usepackage{mathtools}
\newcommand{\smashtemp}[1]{\smash{#1}}

\newcommand{\vect}[1]{\smashtemp{\vec{#1}}}

\newcommand{\smallnorm}[1]{\smashtemp{\left\|\kern-1pt #1\kern-1pt \right\|}}

\newcommand{\TransitionMatrix}{M}

\newcommand{\mixingTimeM}{\tau_{\!\TransitionMatrix}^{ }}

\newcommand{\mixingTimeER}{\tau_{\!M_{\textrm{ER}}}^{ }}

\newcommand{\betag}{\beta_{g}^{ }}
\newcommand{\betagTextNotSmash}{\mbox{$\betag$}}
\newcommand{\betagText}{\smash{\betagTextNotSmash}}

\newcommand{\mug}{\mu_{g}^{ }}
\newcommand{\mugTextNotSmash}{\mbox{$\mug$}}
\newcommand{\mugText}{\smash{\mugTextNotSmash}}

\newcommand{\kappag}{\kappa_{g}^{ }}
\newcommand{\kappagTextNotSmash}{\mbox{$\kappag$}}
\newcommand{\kappagText}{\smash{\kappagTextNotSmash}}

\newcommand{\muEdge}{\mu_{\EdgeSub}^{ }}
\newcommand{\muEdgeTextNotSmash}{\mbox{$\muEdge$}}
\newcommand{\muEdgeText}{\smash{\muEdgeTextNotSmash}}

\newcommand{\muWedge}{\mu_{\WedgeSub}^{ }}
\newcommand{\muWedgeTextNotSmash}{\mbox{$\muWedge$}}
\newcommand{\muWedgeText}{\smash{\muWedgeTextNotSmash}}

\newcommand{\kappaWedge}{\kappa_{\WedgeSub}^{ }}
\newcommand{\kappaWedgeTextNotSmash}{\mbox{$\kappaWedge$}}
\newcommand{\kappaWedgeText}{\smash{\kappaWedgeTextNotSmash}}

\newcommand{\kappaTriangle}{\kappa_{\TriangleSub}^{ }}
\newcommand{\kappaTriangleTextNotSmash}{\mbox{$\kappaTriangle$}}
\newcommand{\kappaTriangleText}{\smash{\kappaTriangleTextNotSmash}}

\newcommand{\scaledCum}{\kappa_{g}^{ }/\mu^r_{\EdgeSub}}
\newcommand{\scaledCumTextNotSmash}{\mbox{$\scaledCum$}}
\newcommand{\scaledCumText}{\smash{\scaledCumTextNotSmash}}

\newcommand{\scaledCumWedge}{\kappa_{\WedgeSub}^{ }/\mu_{\EdgeSub}^{2}}
\newcommand{\scaledCumWedgeTextNotSmash}{\mbox{$\scaledCumWedge$}}
\newcommand{\scaledCumWedgeText}{\smash{\scaledCumWedgeTextNotSmash}}

\newcommand{\scaledCumTriangle}{\kappa_{\TriangleSub}^{ }/\mu_{\EdgeSub}^{3}}
\newcommand{\scaledCumTriangleTextNotSmash}{\mbox{$\scaledCumTriangle$}}
\newcommand{\scaledCumTriangleText}{\smash{\scaledCumTriangleTextNotSmash}}

\newcommand{\scaledCumSquare}{\kappa_{\SquareSub}^{ }/\mu_{\EdgeSub}^{4}}
\newcommand{\scaledCumSquareTextNotSmash}{\mbox{$\scaledCumSquare$}}
\newcommand{\scaledCumSquareText}{\smash{\scaledCumSquareTextNotSmash}}

\newcommand{\ERrho}{\textrm{ER}_{n,\rho}^{ }}
\newcommand{\ERrhof}{\mbox{$\ERrho$}}

\newcommand{\ERrhohalf}{\textrm{ER}_{n,\nicefrac{1}{2}}}
\newcommand{\ERrhohalff}{\mbox{$\ERrhohalf$}}
\newcommand{\ERrhohalfTextNotSmash}{\ERrhohalff}
\newcommand{\ERrhohalfText}{\smash{\ERrhohalfTextNotSmash}}

\newcommand{\ERmu}{\textrm{ER}_{n,\mu_{\EdgeSub}}}
\newcommand{\ERmuTextNotSmash}{$\ERmu$}
\newcommand{\ERmuText}{\smash{\ERmuTextNotSmash}}

\renewcommand{\footnotesize}{\small}

\renewcommand{\labelenumi}{\textit{\theenumi}}
\renewcommand{\theenumi}{\textcolor{Blue}{\textit{\arabic{enumi}}.}}

\newcommand{\formatInstructionNumberQuestions}[1]{\textcolor{black}{\textit{#1}.}}

\usepackage{hyperref}
\usepackage{comment}

\usepackage[accepted]{icml2023}

\usepackage{amsmath}
\usepackage{amssymb}
\usepackage{mathtools}
\usepackage{amsthm}

\usepackage[capitalize,noabbrev]{cleveref}

\theoremstyle{plain}

\theoremstyle{definition}

\theoremstyle{remark}

\usepackage{footmisc}

\icmltitlerunning{Quantifying Human Priors over Social and Navigation Networks}

\begin{document}

\twocolumn[
\icmltitle{Quantifying Human Priors over Social and Navigation Networks}

\begin{icmlauthorlist}
\icmlauthor{Gecia \mbox{Bravo-Hermsdorff}}{y} 
\end{icmlauthorlist}

\icmlaffiliation{y}{Department of Statistical Science, University of London, UK}

\icmlcorrespondingauthor{\\Gecia \mbox{Bravo-Hermsdorff}}{gecia.bravo@gmail.com}

\icmlkeywords{human priors, relational data, graphs, inductive biases, social networks, navigation networks}

\vskip 0.3in
]

\printAffiliationsAndNotice{}

\begin{abstract} 
Human knowledge is largely implicit and \mbox{relational} ---   
do we have a friend in common? can I walk from here to there? 
In this work, we leverage the combinatorial structure of graphs to quantify human priors over such relational data. 
Our experiments focus on two domains that have been continuously relevant over evolutionary timescales: social interaction and spatial navigation.  
We find that some features of the \mbox{inferred} priors are remarkably consistent, 
such as the tendency for sparsity as a function of graph size.  
Other features are \mbox{domain-specific}, 
such as the propensity for triadic closure in social interactions.  
More broadly, our work demonstrates how nonclassical statistical analysis of indirect \mbox{behavioral} experiments can be used to efficiently model latent biases in the data.
\end{abstract}

\section{Brains Rely on Efficient Priors}
\label{sec:efficientcoding} 
A foundational result in the fields of artificial intelligence, neuroscience, and psychology is the establishing of the central role played by inductive biases or priors in learning 
\cite{shiffrin2020brain, wolpert2021important}. 
The importance of priors cannot be understated; indeed, their quantification elucidates many aspects of our perception and cognition.

\ParagraphVertSpace
\textbf{The efficient coding hypothesis. } 
Examples of how \mbox{priors} inform neuroscience can often be understood through the lens of the ``efficient coding hypothesis'', 
which states that neural representations 
have adapted to \mbox{efficiently} encode the relevant statistics of our environment \cite{barlow1961possible}. 
%

Several decades of work investigating and refining this hypothesis have contributed to major advances in our understanding of the neural code \cite{manookin2023two}. 
For example, many properties of mammalian visual cells (such as sensitivity to orientation and spatial frequency) have been shown to be optimized for transmitting information about natural scenes \cite{simoncelli2003vision,field1989what}.  

Likewise, the mammalian cochlea and auditory nerve fibers have properties that allow for efficient representation of the acoustic structure of speech and other natural sounds \cite{lewicki2002efficient,mcdermott2013summary}. 
Other computations, such as visual attention \cite{orban2008bayesian} and working memory \cite{mathy2012s,brady2009compression}, display analogous properties. 

\ParagraphVertSpace
\textbf{Visual priors color our perception. }
This principle of efficient coding also explains several \mbox{well-known} visual illusions \cite{howe2005themuller, howe2005thepoggendorff}, 
evidentiating a general and important computational tradeoff in biology: 
priors cannot be both exhaustive and efficient. 
That is, by efficiently distinguishing relevant visual information, 
our priors render us blind to insignificant differences.  
Might there be similar ``illusions'' with respect to our priors over the structure of connections?
\section{Tasks are Often Relational}
\label{sec:motivation}
From roads between places, websites on the internet, words in a text, and friendships between people, humans are routinely confronted with a web of things (nodes) structured in terms of their relations (edges). 
Despite the pervasiveness of networks in our lives, 
knowledge of our priors about them is rather sparse. 

However, one notable paradigm is the learning of network structure from random walks \cite{lynn2020humans, klishin2022exposure}.  
This approach typically consists of showing nodes to participants in a temporal sequence that respects the structure of the network (such as samples of random walks),   
and comparing the ease with which they learn these transitions for several different networks \cite{schapiro2013neural, tompson2019individual}. 

Complementing these detailed experiments on specific networks, 
our work focuses on quantifying humans' initial beliefs about all such networks. 
The motivating question is: 
given a set of $n$ things and minimal or no information about how they relate, 
what is the prior likelihood assigned to each of the many possible patterns of connections? 

\ParagraphVertSpace
\textbf{Why we focus on navigation and social networks. }
Tasks related to spatial navigation and social interaction have been quotidian over evolutionary timescales. 
Thus, our brains have likely adapted to efficiently encode them. 
Indeed, there is much evidence supporting this hypothesis. 
For example, the hippocampus encodes a spatial map of the environment \cite{maguire2003navigation,eichenbaum2017role}. 
Likewise, there are brain regions specialized in the processing of social information and theory of mind (i.e., the modeling of others' mental states) \cite{richardson2019development,devaine2014social}. 

Additionally, these two domains are qualitatively different: 
spatial navigation networks are constrained by physical space, 
whereas social networks are more abstract and interconnected. 
Comparing the similarities and differences of our priors in these domains could aid in building a more complete understanding of their associated neural processes.

\ParagraphVertSpace
\textbf{Key contributions. } 
\finalSolve{We develop a framework for quantifying human priors over relational data (sections~\ref{sec:Framework} and \ref{sec:MCMCP}), 
and summarize the results in a meaningful way (section~\ref{sec:Analysis}).} 
\section{Overview of our Framework}
\label{sec:Framework}
The number of unique configurations of connections between $n$ nodes grows superexponentially\footnote{For a feeling for the scaling, see table~\ref{table:numgraphs} (appendix~\ref{app:scalability}).} \citep{oeis}.  %
This poses several difficulties in quantifying human priors over such graphs: 
\vspace{-7pt}
\begin{enumerate}
    \item engaging human attention in experiments that involve reasoning about such a large number of possibilities; \\\vspace{-15pt}
    \item properly sampling the space of graphs; and \\\vspace{-15pt}
    \item meaningfully summarizing and comparing priors over such a high-dimensional space. \\\vspace{-15pt}
\end{enumerate}
\vspace{-5pt}
We now provide a brief overview of how our framework overcomes these challenges, 
\finalSolve{expanding on the details in the subsequent sections.}
\subsection{Engaging Human Attention}
\label{ssec:HumanAttention}
\begin{figure*}[hbt!]
	\begin{center}
	\includegraphics[trim={0cm 0cm 0.03cm 0cm},clip,width=1\columnwidth]{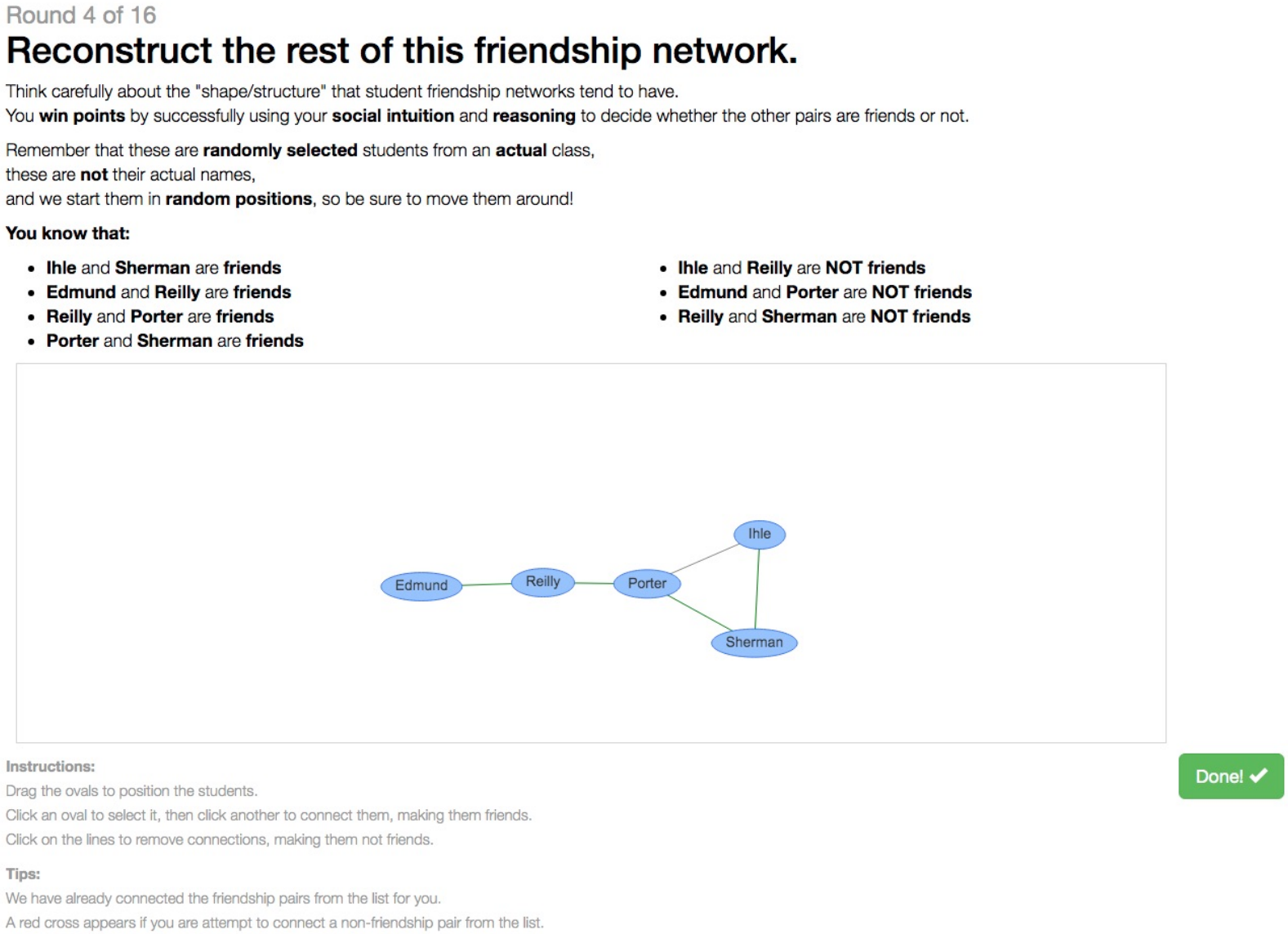}
	\hfill
	\includegraphics[trim={0cm 0cm 0.00cm 0cm},clip,width=0.95\columnwidth]{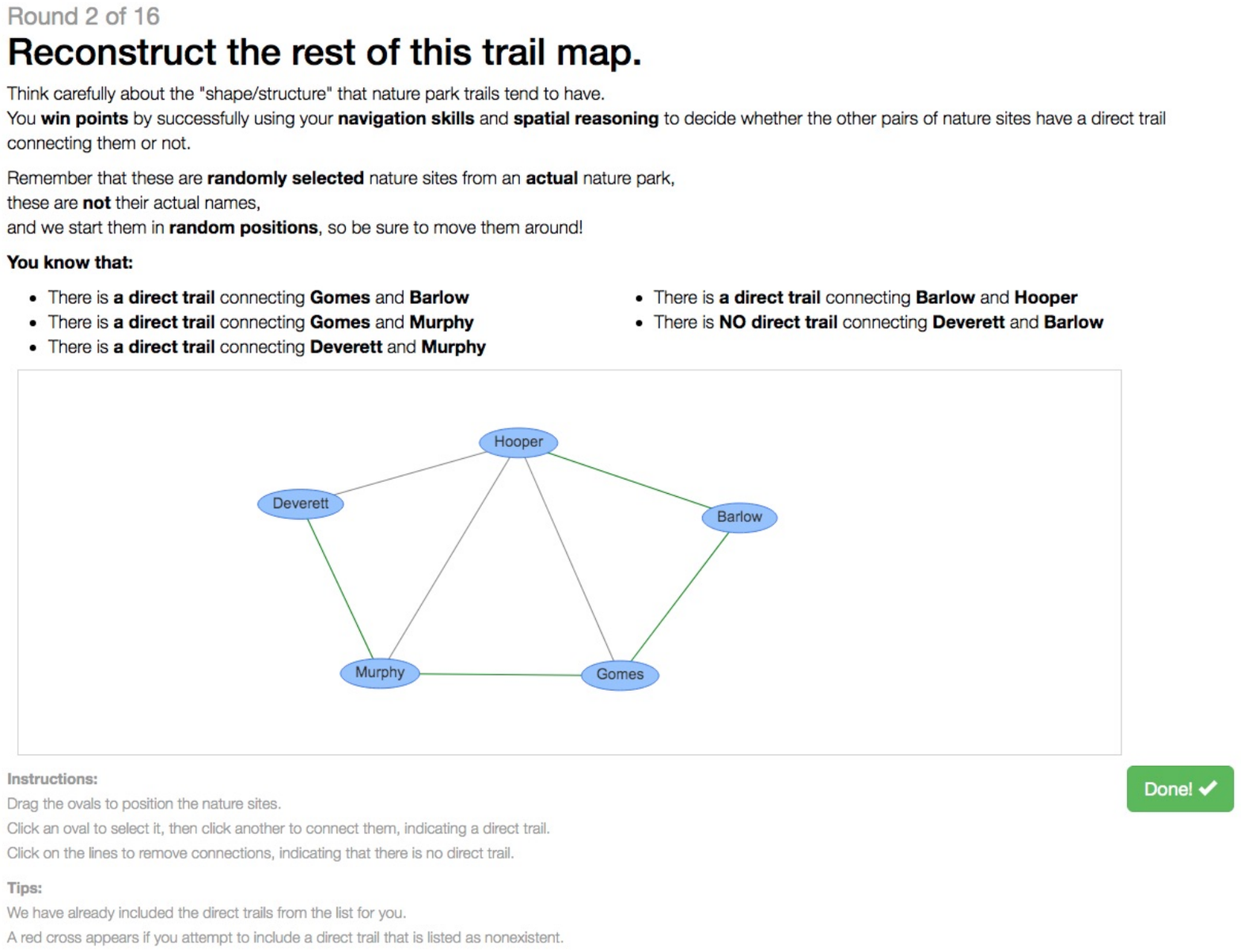}
	\vspace{-3pt}
  \caption{
    \textbf{Screenshots of our main experimental interface}  \textbf{for two cover stories:} \textbf{social \classcolor{class}} \textit{(left)}  \textbf{and} \textbf{navigation \trailcolor{park}} \textit{(right)}.\\ 
Our online platform allowed participants to easily ``draw'' their inferences about the obscured relations of a graph 
(demo \href{https://www.youtube.com/watch?v=aZNeN293MZs}{\underline{video}}).
Note that the two images above are nearly identical:  
to make the comparisons as fair as possible, 
we made the experiments identical in every aspect, except for the text \textit{specifically} related to each cover story.
See appendix~\ref{app:experimentalprocedure} for a detailed description of these experiments and \mbox{high-resolution} versions of these images (figs.~\ref{fig:ScreenshotSocStudent} \classcolor{class} and~\ref{fig:ScreenshotNavTrail} \trailcolor{park}).
}
  \label{fig:ScreenshotExperiments}
	 \end{center}	
\end{figure*}

We built an online experimental platform that 
allowed participants to easily draw their inferences about obscured relations in a graph (demo \href{https://www.youtube.com/watch?v=aZNeN293MZs}{\underline{video}} and  fig.~\ref{fig:ScreenshotExperiments}). 
In brief, participants were shown ``partial graphs'', 
containing all the nodes, but only some of the pairwise relations between them, and were then asked to infer the existence (or absence) of the remaining relations.  
The meaning of these graphs was given by one of the four cover stories that served as the context for our experiments (table~\ref{table:coverstories}). 

Deploying this platform in MTurk \cite{turk},
we carefully curated a large amount of human data in experiments involving social and spatial navigation networks (over $1200$ participants and $15000$ data points). 
To ensure that the final design was as engaging and intuitive as possible, we ran a variety of pilot experiments (over $300$ participants).  
This effort paid off: 
for the final experiments, 
the \mbox{post-questionnaire} feedback was quite positive (several MTurk workers even sent personal emails about how enjoyable the ``game'' was!), 
and the data were of high quality (see appendix~\ref{app:exclusioncriteria} for details).
\vspace{-14pt}
\begin{table}[ht]
\caption{\textbf{Content of the four cover stories of our experiments.} 
Participants were asked to infer the presence or absence of\\obscured \tableFontTop{relations} between pairs of \tableFontTop{nodes} in a \tableFontTop{context}. 
}
\vskip 0.05in
{\renewcommand{\arraystretch}{1.4}
\begin{tabular}{c|c|c|c}
    \tableFontTop{domain}                     &   \tableFontTop{context}    &  \tableFontTop{nodes}      &   \tableFontTop{relations}\\
    \hline
    \multirow{2}{*}{\tableFontSocNav{social}}     &   \classcolor{\tableFontStoryName{class}}     &   \classcolor{\tableFontEntries{students}}    &  \classcolor{\tableFontEntries{friendships}}\\
    \cline{2-4}
                                &   \workcolor{\tableFontStoryName{work}}     &   \workcolor{\tableFontEntries{coworkers}}   &   \workcolor{\tableFontEntries{friendships}}\\
        \hline
    \multirow{2}{*}{\tableFontSocNav{navigation}} &   \citycolor{\tableFontStoryName{city}}   &   \citycolor{\tableFontEntries{neighborhoods}}     &   \citycolor{\tableFontEntries{borders}}\\
    \cline{2-4}
                                &   \trailcolor{\tableFontStoryName{park}}          &   \trailcolor{\tableFontEntries{nature sites}}    &   \trailcolor{\tableFontEntries{trails}}\\
\end{tabular}
}
\label{table:coverstories}
\end{table}

\vspace{-16pt}
\subsection{Sampling the Space of Graphs}
\label{sse:properlysampling}
For the first participants, we initialized our experiments using graphs spanning a wide range of edge densities.  
From these, we generated partial graphs, and asked participants to infer the remaining relations. 
We then repeatedly used the responses from the previous participants to generate partial graphs for the next participants (fig.~\ref{fig:MCMCPschemeEight}). 
In essence, our online platform instantiates a ``Markov Chain Monte Carlo algorithm with People (MCMCP)'', 
with multiple chains being built in parallel. 

Standard implementations of MCMCP experiments model the (shared) prior of the participants by sampling (some of) their responses from (sufficiently long) experimental chains. 
Here, we use the data more efficiently (figs.~\ref{fig:FigFitBoth} and~\ref{fig:ChainLengthFixed}) by leveraging the graphical structure to fit the aggregated responses of participants to a natural Bayesian model (section~\ref{sec:BayesianAnsatz}). 
In particular, we parameterize their priors using a hierarchical family of maximum entropy distributions over graphs (sections~\ref{sec:maximumentropypriors} and~\ref{sec:modelselection}), which offer ``smooth''\footnote{In the sense that graphs that differ by fewer edges are assigned similar probabilities.} \mbox{low-dimensional} parameterizations of the \mbox{high-dimensional} space of graphs (figs.~\ref{fig:RecoverSyntheticData} and~\ref{fig:GeneralizationInData}). 
\subsection{Summarizing and Interpreting Priors over Graphs}
\label{ssec:SumInterPriorOverGraphs}
\finalSolve{Graph cumulants \cite{bravo2021principled, gunderson2020introducing} capture what is typically meant by ``substructure'' or  ``motif'': a subgraph $g$ whose prevalence in a distribution over graphs is statistically different from that which would be expected due to the prevalence of smaller subgraphs contained in $g$. 
Here, we compare the inferred priors using the scaled version of graph cumulants, 
which additionally takes into account the density of connections (figs.~\ref{fig:egalitarian} and \ref{fig:socialtriangles}).} 
%

\section{Experimental Design} 
\label{sec:MCMCP}
In this section, we provide a brief overview of the literature on MCMC with People (sections~\ref{subsec:MCMCPhist} and \ref{sec:MCMCPmodel}), 
the general framework to which our method can be applied.  
We then describe our experiments (sections~\ref{sec:MCMCPovergraph} and~\ref{sec:coverstoryandtable}). 
\subsection{Markov Chain Monte Carlo with People (MCMCP)}
\label{subsec:MCMCP}
\subsubsection{Related Work}
\label{subsec:MCMCPhist}
Iterated learning refers to the process whereby a participant learns from data generated by another participant, 
who themselves learned it the same way, and so on. 
It is a \mbox{highly-researched} and ubiquitous psychological phenomenon --- language and cultural evolution being two important examples \cite{kirby2014iterated, morgan2020experimental}. 
Under some assumptions (see appendix~\ref{app:mcmcpappendix}), 
iterated learning can be modelled as a Markov Chain Monte Carlo algorithm instantiated by the Participants (algorithm~\ref{alg:GenericMCMCPExperiment}) that has as its stationary distribution their shared prior over the relevant space \cite{grifiths2007language}.
\vspace{-0pt}
\begin{algorithm}[h]
    \caption{\textsc{Generic MCMCP Experiment}}
    \label{alg:GenericMCMCPExperiment}
    \begin{algorithmic}
        \STATE \textbf{Initialize:}  {\InfoOutWord}$_0^{ }$
        \FOR{$t=1$ {\bfseries to} $T$} 
        \STATE {\InfoInWord}$_t^{ }$ $ = $ \experimentalist$(${\InfoOutWord}$_{t-1}^{ }$$)$
        \STATE {\InfoOutWord}$_t^{ }$ $ = $ \participant$_t^{ }$$(${\InfoInWord}$_{t}^{ }$$)$
        \ENDFOR
    \end{algorithmic}
\end{algorithm}
\vspace{-5pt}

This MCMCP model has been employed to quantify human priors in a variety of contexts, such as: 
locations in visual scenes \cite{langlois2021serial};  
variations in facial features \cite{uddenberg2018teleface};  
strengths of causal relationships \cite{yeung2015identifying};  
moral categories of words in ethics \cite{hsu2019identifying}; 
names of colors \cite{xu2013cultural}; 
and 
kernels for Gaussian processes \cite{schulz2017compositional}. 
This framework has also been applied to quantify priors of (\mbox{non-human}) large language models \cite{yamakoshi2022probing,marjieh2022analyzing}. 
\subsubsection{The MCMCP Model}
\label{sec:MCMCPmodel}
Let \M{\InfoIn} be the space of all combinations of {\InfoInWordPCap}
participants might be given in the experiment. 
%
And let \M{\InfoOut} be the space of all {\InfoOutWordPCap} that the participants might consider when giving their responses. 
We assume that both spaces are discrete and finite, and denote the space of probability distributions over them as \M{P(\InfoIn)} and \M{P(\InfoOut)}, respectively.

The \textsc{Experimentalist} uses the {\InfoOutWord} of the previous participant (i.e., their response) to generate noisy/partial {\InfoInWord} to give to the next participant.  
This process is a probabilistic map from $\InfoOut\!\rightarrow\!\InfoIn$, denoted in algorithm~\ref{alg:GenericMCMCPExperiment} as 
\mbox{\experimentalist$(\cdot)$}, 
with associated probability distribution \mbox{\M{p(\InfoInOne|\InfoOutOne)}}. 

%
Given a prior distribution \mbox{\M{\Prior\in P(\InfoOut)}}, and presented with {\InfoInWord} \mbox{\M{\InfoInOne\in\InfoIn}}, 
a ``Bayesian'' \mbox{\textsc{Participant}} will sample a {\InfoOutWord} \mbox{\M{\InfoOutOne\in\InfoOut}} from their posterior distribution as their response:
\vspace{-4pt}
\begin{equation*}
  p(\InfoOutOne|\InfoInOne) = \frac{p(\InfoInOne|\InfoOutOne)\Prior(\InfoOutOne)}{\sum_{\InfoOutOne \in \InfoOut}p(\InfoInOne|\InfoOutOne)\Prior(\InfoOutOne)} 
\end{equation*}
This process is a probabilistic map from $\InfoIn\!\rightarrow\!\InfoOut$, and is denoted in algorithm~\ref{alg:GenericMCMCPExperiment} as 
\participant$(\cdot)$. 


If all participants have a shared prior distribution \M{\Prior} over the {\InfoOutWordPCap},
the composition of these stochastic maps 
\mbox{\M{\ParticipantMapping(\ExpMapping(\cdot))\!:\!P(\InfoOut)\!\rightarrow\!P(\InfoOut)}} 
has this prior \M{\Prior} as its unique stationary distribution (given the standard technical conditions on the Markov chain, see appendix~\ref{app:mcmcpassumptions}). 
\subsubsection{Relational MCMCP}
\label{sec:MCMCPovergraph}
\begin{figure}[H]
	\begin{center}		\centerline{\includegraphics[width=1\columnwidth]{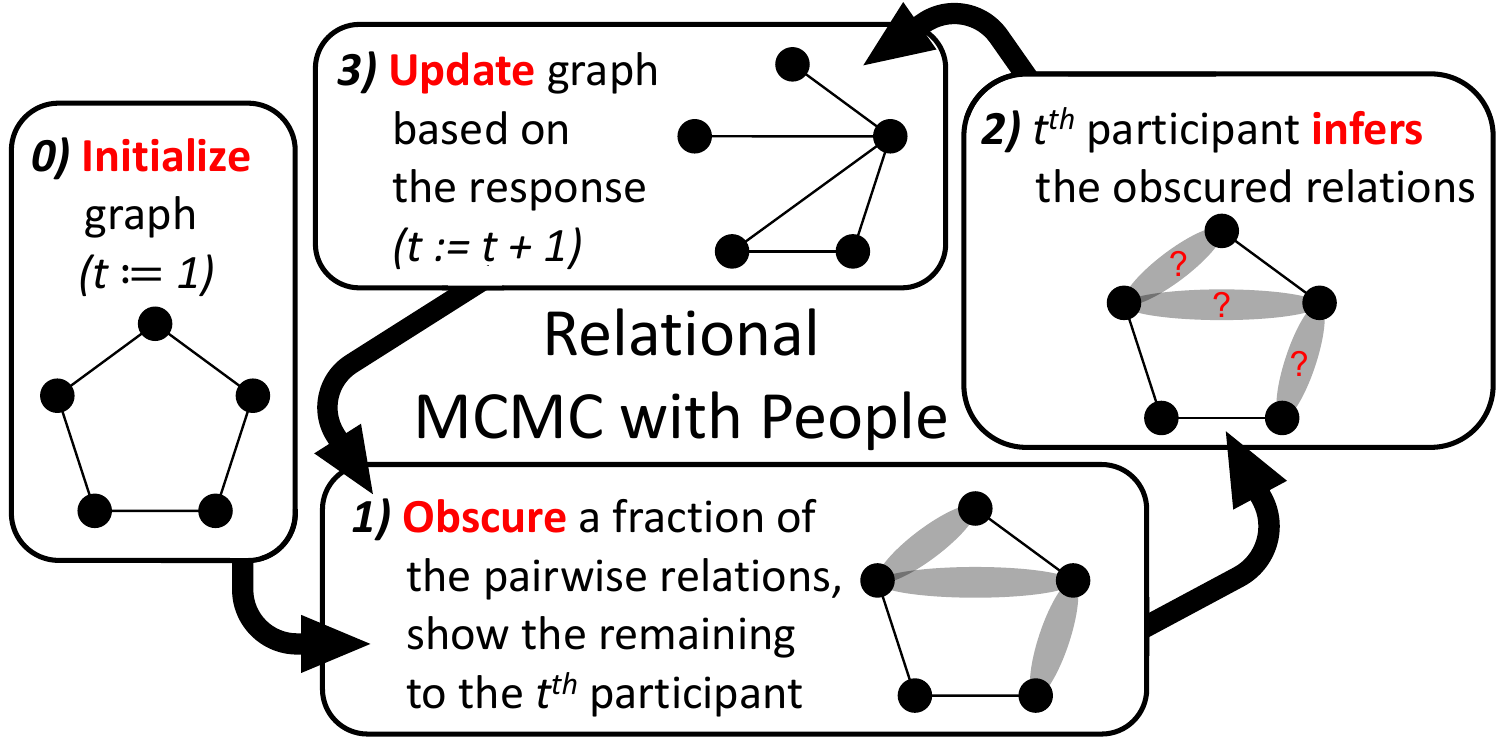}}
  \vspace{-3pt}
  \caption{\textbf{Algorithm for generating a round of our experiment.} 
          The context of the experiment was given by one of the four cover stories in table~\ref{table:coverstories}, and an interface allowed participants to easily manipulate the graphs (see demo \href{https://www.youtube.com/watch?v=aZNeN293MZs}{\underline{video}} and fig.~\ref{fig:ScreenshotExperiments}).
  Each participant did multiple rounds, corresponding to different chains. 
  }
  \label{fig:MCMCPschemeEight}
	\end{center}	
\end{figure}

Figure~\ref{fig:MCMCPschemeEight} illustrates our algorithm for generating MCMCP experiments on graphs. 
It consists of the following steps:
\vspace{-4pt} 
\begin{enumerate}\addtocounter{enumi}{-1}
    \item \textit{Initialize} the chain with a graph containing $n$ nodes and $\binom{n}{2}$ pairwise relations between them (e.g., the friendships, or lack thereof, between students in a \classcolor{class}). \\\vspace{-16pt}
    \item \textit{Obscure} a random fraction $\fractionObscure$
    of this graph's pairwise relations (e.g., \mbox{$\fractionObscure=\nicefrac{3}{10}$} in fig.~\ref{fig:MCMCPschemeEight}). 
    \\\vspace{-16pt}
    \item Based on this ``partial graph'', ask the participant to \textit{infer} the obscured relations.\\\vspace{-16pt} 
    \item \textit{Update} the graph based on their response.\label{step3RelationalMCMCP}\\\vspace{-16pt}
     \item \textit{Repeat} the sequence of steps \textit{\textcolor{Blue}{1}}, \textit{\textcolor{Blue}{2}}, and \textit{\textcolor{Blue}{3}}, each time with a new participant.\\\vspace{-18pt}
\end{enumerate}
Our experiments focused on simple graphs: undirected unweighted graphs with no \mbox{self-loops} or parallel edges.\footnote{So, a simple graph {\GraphText} with $n$ nodes has \mbox{$\binom{n}{2}$} pairwise relations.} 

For a given chain of our experiment, the space of {\InfoOutWordPCap} is 
\mbox{\M{\GraphSet_{n}^{ }}}, the set of simple graphs with $n$
nodes, 
and \mbox{$\Graph_t^{ } \in \GraphSet_{n}^{ }$} denotes the response of the participant in the $t^{\text{th}}$ iteration/round.  
The space of {\InfoInWordPCap} is \smash{\mbox{\M{\PartialGraphSet_{n,\numObs}^{ }}}}, the set of all partial graphs with $n$ nodes and $\numObs$ of the 
pairwise relations obscured, 
and \mbox{$\PartialGraph_t^{ } \in \PartialGraphSet$} denotes the specific partial graph shown to the participant in the $t^{\text{th}}$ iteration/round. 
%
The map \smash{\mbox{\M{\ExpMapping:\GraphSet_n^{ }\rightarrow\PartialGraphSet_{n,\numObs}^{ }}}} takes a graph on $n$ nodes, and makes a partial graph by randomly obscuring $\numObs$  pairwise relations. 
\subsection{Experimental Platform and Cover Stories}
\label{sec:coverstoryandtable}
We developed an online experimental platform and recruited participants using MTurk.  
Our platform allows participants to easily draw their inferences about the obscured relations in the graph (demo \href{https://www.youtube.com/watch?v=aZNeN293MZs}{\underline{video}}) and allocates them to one of multiple experimental chains in real time. 
%
Each experiment had one of four different cover stories:  
two in the social domain and two in the navigation domain (table~\ref{table:coverstories}).

In a given experiment, a participant gave responses for many rounds, where each round was part of a different chain. 
A response was included in a chain (and thereby used to generate the partial graph for the next participant) only if it passed \mbox{judiciously-chosen} criteria. 
In appendix~\ref{app:experimentalprocedure}, we provide a more detailed description of the experiments, the data collection, and the data cleaning procedures. 

The experiment begins with an introduction about the particular cover story and poses several questions to the participant to ensure their understanding.  
After a video demonstration of the interactive platform, each round begins with the partial graph at the center of the interactive interface and
a list of the ``unobscured'' relations at the top of the screen. 
Using this interface, the participant were able to move the nodes and add/remove edges (fig.~\ref{fig:ScreenshotExperiments}).  
Once the participant submitted their response, a question appears about which node(s) they thought to be the most/least important (asked in a variety of ways). 
To incentivize participants to respond using their true prior, 
they were told during the introduction that there is a ground truth and they would be rewarded for correctly guessing the relations that were obscured (see appendix~\ref{app:screenshots} for the complete text). 
Clearly, as there is \textit{not} such a ground truth, their responses did not influence their final payoff (though their level of engagement did, see appendix~\ref{app:exclusioncriteria}). 

As the analysis assumes that the nodes are exchangeable, we aimed to make their labels as neutral as possible.  
To this end, we randomly selected the node labels from a long list of last names, 
while ensuring that no name was repeated during an experiment. 
Participants were clearly instructed that the node labels were fictitious names, and that they provided no information about the ``correct'' answers.  
\section{Data Analysis}
\label{sec:Analysis}
In this section, we describe how we analyze the data from our experiments.
We first discuss how data from MCMCP experiments is typically used to obtain priors and some of the limitations of this standard approach. 
We then introduce our method, which alleviates these issues 
by exploiting the Bayesian assumption (section~\ref{sec:BayesianAnsatz}), and uses the combinatorial structure of graphs to fit the data with a natural family of maximum entropy models (sections~\ref{sec:maximumentropypriors}, and \ref{sec:modelselection}).  
We end this section by describing how we quantify and compare relevant characteristics of the inferred priors (sections~\ref{sec:graphcumulants}, \ref{sec:scaledcumulants}, 
and~\ref{sec:errorbarplots}).
\subsection{Limitations of the Standard MCMCP Approach}
\label{sec:samplingapproach}
Typically, studies employing MCMCP experiments 
use the participants' responses towards the end of the chains as a proxy for their prior.  
Indeed, according to the assumptions of the MCMCP model, the stationary distribution of (Bayesian) participants' responses 
approximates their (shared) prior over the relevant {\InfoOutWordPCap}.

However, this approach wastes much of the collected data for two reasons. 
First, one must discard the initial responses until the chain has (hopefully\footnote{Determining convergence can be highly nontrivial in certain cases, especially when the state space is large \cite{roy2020convergence}.})
converged (sufficiently close\footnote{This can also be nontrivial to determine.}) to its stationary distribution, 
the \mbox{so-called} \mbox{``burn-in''} period \cite{raftery1996implementing}. 
Second, as the responses are correlated, one cannot treat them as completely independent samples, and thus has fewer effective samples \cite{hsu2015mixing}. 
Moreover, the number of iterations/rounds required for an experimental chain to be sufficiently close to its stationary distribution (i.e., its mixing time) is highly dependent on the probabilistic mapping \mbox{\M{\ExpMapping\!:\!\InfoOut\!\rightarrow\!\InfoIn}}  used to generate the experiments, and on the participants' (unknown) prior \M{\Prior} (see figs.~\ref{fig:ERuniversalCurve} and~\ref{fig:ConvergenceToPrior}).

While this might not always be a problem (such as when samples can be efficiently generated by a computer),
using human participants to generate samples in MCMCP often presents a significant bottleneck (controlling electrons is typically simpler than controlling human attention). 
\subsection{Leveraging the Bayesian Ansatz}
\label{sec:BayesianAnsatz}
Instead of using (a few of) the observed responses \M{\InfoOutOne_t} to approximate the prior, we exploit the fact that the paired data \smash{\M{(\InfoInOne_t\rightarrow \InfoOutOne_t)}} are more informative than these \M{\InfoOutOne_t} alone.
Specifically, as the traditional approach already assumes that the participants are Bayesian with the same prior, we make explicit use of this implicit assumption by modelling the paired data \smash{\M{(\InfoInOne_t\rightarrow \InfoOutOne_t)}} in terms of the transition matrix \smash{\mbox{$p(\InfoInOne|\InfoOutOne)$}} induced by an underlying shared prior $\Prior$.  
In appendix~\ref{app:mcmcpbayesian}, we show that this fitting approach alleviates typical problems of standard MCMCP data analysis: estimating the mixing time  (fig.~\ref{fig:FigFitBoth}) and correlated samples (fig.~\ref{fig:ChainLengthFixed}). 

An important hurdle still remains:
for this fitting approach to work, one must be able to parameterize the priors.  
However, 
as mentioned in section~\ref{sec:Framework},
the number of nonisomorphic graphs grows superexponentially in their number of nodes \M{n}.
It is thus unwise to fit a multinomial distribution to each of these graphs, thereby, effectively treating them as incomparable variables. 
%
%
To obtain informative priors, 
we must use a meaningful \mbox{low-dimensional} smooth 
parameterization of probability distributions over graphs, 
such that graphs that differ by fewer edges 
are given similar probabilities. 
%

We now describe a natural hierarchical family of network models that provides such a parameterization (section~\ref{sec:maximumentropypriors}) and
how we fit these models to the data (section~\ref{sec:modelselection}).
In appendix~\ref{sec:priorparameterization}, we show that this choice of parameterization for the priors leads to a more accurate recovery of the prior in simulated data (where the ground truth is known) (fig.~\ref{fig:RecoverSyntheticData}) and improved generalization in real data (fig.~\ref{fig:GeneralizationInData}).
\subsection{Modeling the Priors using a Hierarchy of\\Maximum Entropy Distributions over Graphs}
\label{sec:maximumentropypriors}
Intuitively, given a set of constraints, the maximum entropy distribution is the ``simplest'' of those that satisfy the constraints \cite{jaynes1957information1, jaynes1957information2}. 
Maximum entropy distributions appear in every corner of science; 
from uniform to Gaussian, beta to binomial, gamma to Poisson and more, nearly all named distributions maximize entropy in some sense.

For simple graphs with $n$ nodes, the simplest statistic is the edge density {\muEdgeText}.  
The maximum entropy distribution corresponding to this statistic is the \mbox{Erd\H{o}s--R\'enyi} model {\ERmuText}, in which a connection appears between each of the \smash{\mbox{${\binom{n}{2}}$}} pairs of the nodes independently with the same probability {\muEdgeText}.
%
In particular, the {\ERrhohalff} distribution assigns uniform probability to each of the graphs with $n$ \textit{labeled} nodes, and serves as the base measure for any maximum entropy distribution over simple graphs with $n$ nodes (\mbox{node-labeled} or not). 

These distributions are known as Exponential Random Graphs Models (ERGMs).  
They have been extensively studied theoretically \cite{chatterjee2013estimating, cimini2019statistical} and applied to a wide range of real networks \cite{lusher2013exponential, lehmann2021characterising}.
Although it is possible to define an ERGM by prescribing any set of realizable constraints, a natural and frequent choice  \cite{lovasz2012large,lauritzen2018random} 
is to prescribe the counts/densities of small subgraphs, such as edges (\raisebox{1pt}{$\Edge$}), ``cherries'' (\raisebox{1pt}{$\Wedge$}), and triangles (\raisebox{1pt}{$\Triangle$}). 

To model the priors, we use the following family of ERGMs:
\vspace{-7pt}
\begin{equation}
    \Prior\big(G\big)\,\propto\,\ERrhohalf\big(G\big) \,\times\, \exp\Bigg\{\sum_{g:E(g)\leq r} \betag \mug\MuShrinkGap\big(G\big) \Bigg\}
    \label{eq:LowDimensionalPrior}
\end{equation}
\vspace{-12pt}

where $G$ is a simple graph with $n$ nodes; 
$\Prior(G)$ is a distribution over such graphs;
 $\mug\MuShrinkGap(G)$ is the (injective homomorphism\footnote{
Consider all injective maps (so no node overlapping) from the nodes of $g$ into the nodes of $G$, $\mug\MuShrinkGap(G)$ is the fraction of such maps for which every edge in $g$ appears in the corresponding location in $G$ (i.e., we do not care about the absence of edges).\label{foot:injhom}\vspace{0pt}})   
density of the subgraph $g$ in the graph $G$; 
{\betagText} is the parameter associated with {\mugText}; 
and \smash{\mbox{$g:E(g)\leq r$}} are all subgraphs with at most $r$ edges. 

That is, this model constrains the densities of all subgraphs with at most $r$ edges (hence the sum over \smash{\mbox{$g:E(g)\leq r$}}).
This choice induces a natural and convenient hierarchy of network models for the priors. 
In particular, the parameter $r$ controls the expressivity of the model. 
For example, \mbox{$r=1$} constrains only the edge density {\muEdgeText}, corresponding to the simplest network model {\ERmuText}. 
Likewise, \mbox{$r=6$} constrains the density of all subgraphs with $6$ edges, leading to a far more complex model for the prior (e.g., this model can exactly specify the probability of all graphs with $4$ nodes). 
\subsection{Fitting and Selecting the Model}
\label{sec:modelselection}
When fitting a model to the prior, we consider all data related to a particular cover story and a given number of nodes.
Specifically, we aggregate data from all such chains regardless of fraction of relations obscured $\fractionObscure$, as when we split the data, there were no significant differences.

For each subgraph $g$ in the model (eq.~\ref{eq:LowDimensionalPrior}), there is a corresponding (conjugate) parameter {\betagText}.  
We fit these parameters numerically by maximizing the \mbox{log-likelihood} of the data:
\vspace{-3pt}
\begin{align*}
\mathcal{L}(\vect{\beta}) = \mathlarger{\sum_t} \log  \frac{\ERrhohalf\kern-1pt\Big(\kern-1pt\Graph_t^{ }\kern1pt\big|\kern1pt\PartialGraph_t^{ }\kern-1pt\Big) \TimesGap \exp\!\Big\{\kern-1pt\vect{\beta}\cdot\vect{\mu}(\Graph_t^{ })\!\Big\}}{\mathlarger{\sum}\limits_{\Graph'\in \GraphSet_{\kern-1pt n}^{ }} \kern-2pt \ERrhohalf\kern-1pt\Big(\kern-1pt\Graph'\kern1pt\big|\kern1pt\PartialGraph_t^{ }\kern-1pt\Big) \TimesGap \exp\!\Big\{\kern-1pt\vect{\beta}\cdot\vect{\mu}(\Graph')\!\Big\}} 
\label{eq:loglikelihooddata}
\end{align*}
\vspace{-10pt}\\
where $G_{t}^{ }$ denotes the participant's response to being shown the partial graph $\PartialGraph_t^{ }$; 
and the distribution $\ERrhohalf\big(\Graph'\kern1pt\big|\kern1pt\PartialGraph_t^{ }\big)$ is that which would be obtained by including edges i.i.d.~with probability $\nicefrac{1}{2}$ for each of the obscured relationships in 
$\PartialGraph_t^{ }$.\footnote{Indeed, this is how participants would reply if their prior \textit{were} {\ERrhohalfText} (corresponding to \mbox{$\vect{\beta}=\vect{0}$}).}  

We employed Newton's method to obtain the parameters {\betagText} for which \smash{\mbox{$\partial \mathcal{L}/\partial \vect{\beta} = 0$}},  
and selected the model complexity $r$
using \mbox{cross-validation} and various sanity checks and robustness tests (see appendix~\ref{app:fittingprocedure}).
\subsection{Interpreting the Data using  Graph Cumulants}
\label{sec:graphcumulants}  
Just as the classical cumulants (e.g., mean, variance, covariance, skew, kurtosis) can be derived from the classical moments, 
so too can graph cumulants be obtained from the subgraph densities (the analogue of moments for graphs \citet{bickel2011method}). 
Graph cumulants are a principled and intuitive family of \mbox{subgraph-based} statistics that naturally captures the propensity (or aversiveness) for any substructure of interest 
(see \citet{bravo2021principled} for a concise application or \citet{gunderson2020introducing} for more details). 

Intuitively, the graph cumulant {\kappagText} quantifies the difference between the observed density {\mugText} of subgraph $g$ and the density that would be expected by chance due to the densities of smaller subgraphs within $g$. 
For example, for the cherry cumulant {\kappaWedgeText}, a term involving the edge density {\muEdgeText} is subtracted from the cherry density:
\smash{\mbox{$\kappaWedge = \muWedge - \mu_{\EdgeSub}^{2}$}}.
For {\kappaTriangleText}, terms involving both the edge and cherry densities, {\muEdgeText} and {\muWedgeText}, are subtracted.
For the simplest random graph {\ERmuText}, 
which has no graphical structure beyond the presence of edges, 
all graph cumulants (aside from {\muEdgeText}\footnote{Just as the mean is the first moment and the first cumulant, the edge density {\muEdgeText} is likewise the first  graph moment {\mugText} and the first  graph cumulant {\kappagText}.}) 
are exactly zero, 
reflecting the fact that larger subgraphs do not require more explanation than just the edge density.
\subsection{Scaling the Graph Cumulants Accounts for Sparsity}
\label{sec:scaledcumulants}
\finalSolve{If one randomly deletes edges i.i.d.~in a graph $G$ such that a fraction $x$ of the edges remain, then the resulting expected edge density {\muEdgeText} is clearly scaled by a factor of $x$ from the original edge density. 
Similarly, for a subgraph $g$ with $r$ edges, its expected subgraph density and graph cumulant are scaled by a factor of \mbox{$x^r_{ }$}.  
This can make comparisons between subgraphs of different sizes difficult, especially for sparse graph distributions.}  
As such, we report the scaled graph cumulants (i.e., {\scaledCumText}) of the inferred priors. 
\subsection{Estimating the Error in our Results}
\label{sec:errorbarplots}
The solid curves in figures~\ref{fig:sparsepriors}, \ref{fig:egalitarian}, and \ref{fig:socialtriangles} display (scaled) graph cumulants of the inferred priors. 
To estimate our uncertainty in these measurements, 
we simulated ideal Bayesian MCMCP agents using the inferred priors, and 
responding to the same partial graphs seen by the participants. 
\finalSolve{We then inferred the prior for each of these synthetic datasets, and computed their (scaled) graph cumulants.}
The shaded regions correspond to \mbox{$\pm 1$}
standard deviation about the average of these values (for $64$ repetitions of this process). 
\section{Results}
\label{sec:ExperimentalResults}
\finalSolve{While our analysis of participants' data makes full use of the MCMCP assumptions (most notably, that participants are Bayesian with the same prior), 
we are not claiming that they exactly hold in practice.  
Nevertheless, the results we present below are remarkably robust (as evidenced by 
model selection, sensitivity analysis, and robustness checks), 
suggesting that the general conclusions are still meaningful.}  
\subsection{Substructures with Noticeable Trends in the Priors}
\label{ssec:resultsedgewedgetriangle}
\begin{figure}[ht]
	\begin{center}
    \centerline{ \includegraphics[trim={0.0cm 0.0cm 0 0},clip,width=0.95\columnwidth]{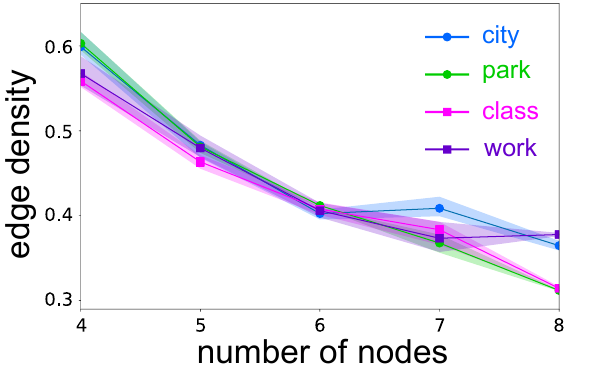}} 
    \vspace{-0.05in}
    \caption{\textbf{Priors over larger graphs have lower edge density.}\\
    Markers in the solid curves correspond to the inferred edge density ({\muEdgeText}) of participants' priors, using the aggregated data of a single cover story with that number of nodes. 
    Shading corresponds to \mbox{$\pm 1$} standard deviation of the average value that would have been obtained if participants all had this inferred prior, and behaved according to the assumptions of the MCMCP model. 
    Note that the result of this procedure is not necessarily centered around the empirical values (i.e., the solid curves). 
} 
 \label{fig:sparsepriors}
 \end{center}
 \vskip -0.3in	    
\end{figure}

\begin{figure}[ht] 
	\begin{center}
 \centerline{ \includegraphics[trim={0.0cm 0.0cm 0 0},clip,width=0.95\columnwidth]{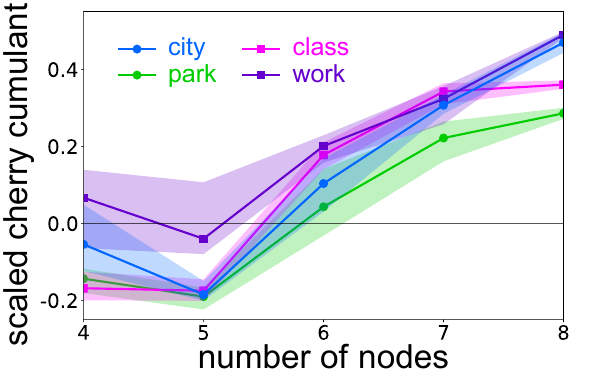}} 
  \vspace{-0.05in}
  \caption{\textbf{Priors over smaller graphs have fewer hubs.}\\
  The analysis is the same as in fig.~\ref{fig:sparsepriors}, but the statistic measured is the scaled cherry cumulant ({\scaledCumWedgeText}), 
  which quantifies preference for degree heterogeneity. 
  A negative value indicates that the prior has edges distributed more uniformly than what would be expected by chance 
  (i.e., in an {\ERmuText} distribution with the same number of nodes $n$ and edge density {\muEdgeText}).}
 \label{fig:egalitarian}
 \end{center}
 \vskip -0.3in	    
\end{figure}

\begin{figure}[ht] 
 \centerline{ \includegraphics[trim={0.0cm 0.0cm 0 0},clip,width=0.95\columnwidth]{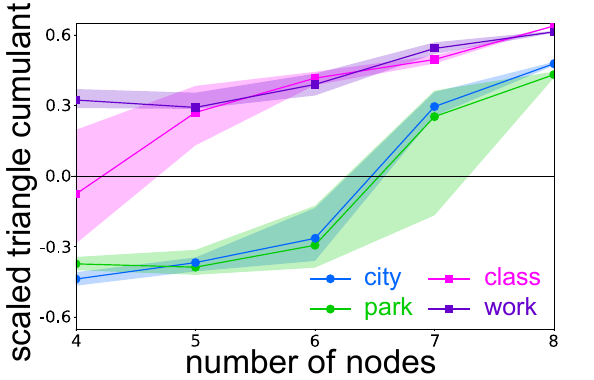}} 
  \vspace{-0.05in}  
  \caption{\textbf{Priors over social graphs have more triangles.}\\
    The analysis is the same as in figs.~\ref{fig:sparsepriors} and~\ref{fig:egalitarian}, but the statistic measured is the scaled triangle cumulant ({\scaledCumTriangleText}),
    which quantifies preference for clustering. 
  A negative value indicates that the prior has fewer triangles than what would be expected by chance. 
  In contrast to figs.~\ref{fig:sparsepriors} and~\ref{fig:egalitarian}, 
 there is a notable difference between the social (\classcolor{class} and \workcolor{work}) 
  and navigation (\citycolor{city} and \trailcolor{park}) domains.} 
 \label{fig:socialtriangles}
\end{figure}

We now present the results for: edge density {\muEdgeText} (fig.~\ref{fig:sparsepriors}), scaled cherry cumulant {\scaledCumWedgeText} 
 (fig.~\ref{fig:egalitarian}), and scaled triangle cumulant {\scaledCumTriangleText}  (fig.~\ref{fig:socialtriangles}) for each of the four cover stories as a function of the number of nodes in the prior.\footnote{Despite number of nodes clearly being a discrete variable, we plot the results as curves to aid in the visualization of the trends.}

Intuitively, these statistics quantify \mbox{well-known} tendencies of real networks: sparsity, degree heterogeneity, and clustering, respectively.
It is perhaps then not a coincidence that the subgraphs associated with these statistics were those that displayed the most discernible trends across the priors. 

\ParagraphVertSpace
\textbf{Priors favor sparsity. } 
As shown in figure~\ref{fig:sparsepriors}, 
we find that the edge density ({\muEdgeText}) systematically decreases as the number of nodes increases.  
The number of connections \textit{per node}, however, appears to be a slowly increasing function of the number of nodes. 
This result is remarkably similar for all the four different cover stories.

\ParagraphVertSpace
\textbf{Priors favor uniform degrees in small graphs. } 
As shown in figure~\ref{fig:egalitarian},
we find that the preference for degree heterogeneity (i.e., a few ``hub'' nodes with many of connections) 
increases as the number of nodes increases. 
The scaled cherry cumulant ({\scaledCumWedgeText})
changes from negative (for graphs with $4$ or $5$ nodes) to positive (for graphs with $6$ or more nodes). 
This suggests that human priors for small graphs favor a notably uniform distribution of node degrees, switching to a preference for heterogeneous node degrees for larger graphs. 
Again, this result is remarkably consistent for all four cover stories. 

\ParagraphVertSpace
\textbf{Priors over social interactions favor triangles. } 
As shown in figure~\ref{fig:socialtriangles},
the scaled triangle cumulant ({\scaledCumTriangleText}) reveals that the priors for the social domain have a notably higher preference for clustering.\footnote{For measuring clustering in bipartite graphs, one should use the scaled \textit{square} cumulant {\scaledCumSquareText}.}
In contrast to the edge (\raisebox{1pt}{$\Edge$}) and cherry (\raisebox{1pt}{$\Wedge$}), this motif (\raisebox{1pt}{$\Triangle$}) clearly distinguishes between the social and navigation domains. 

Indeed, \citet{tompson2019individual} found experimental evidence that humans learn community structure differently when the network is social vs.~\mbox{non-social}. 
Moreover, the prevalence of triadic closure in social networks (i.e., one's friends tend to be friends with each other) is a \mbox{well-established} phenomenon \cite{yang2016social}.
\subsection{Generalization Between and Within Domains}
\label{ssec:generalizationresults}
\begin{figure*}[hbt!]
\begin{center}
  \centerline{\includegraphics[trim={0 0 0 0},clip,width=2.00\columnwidth]{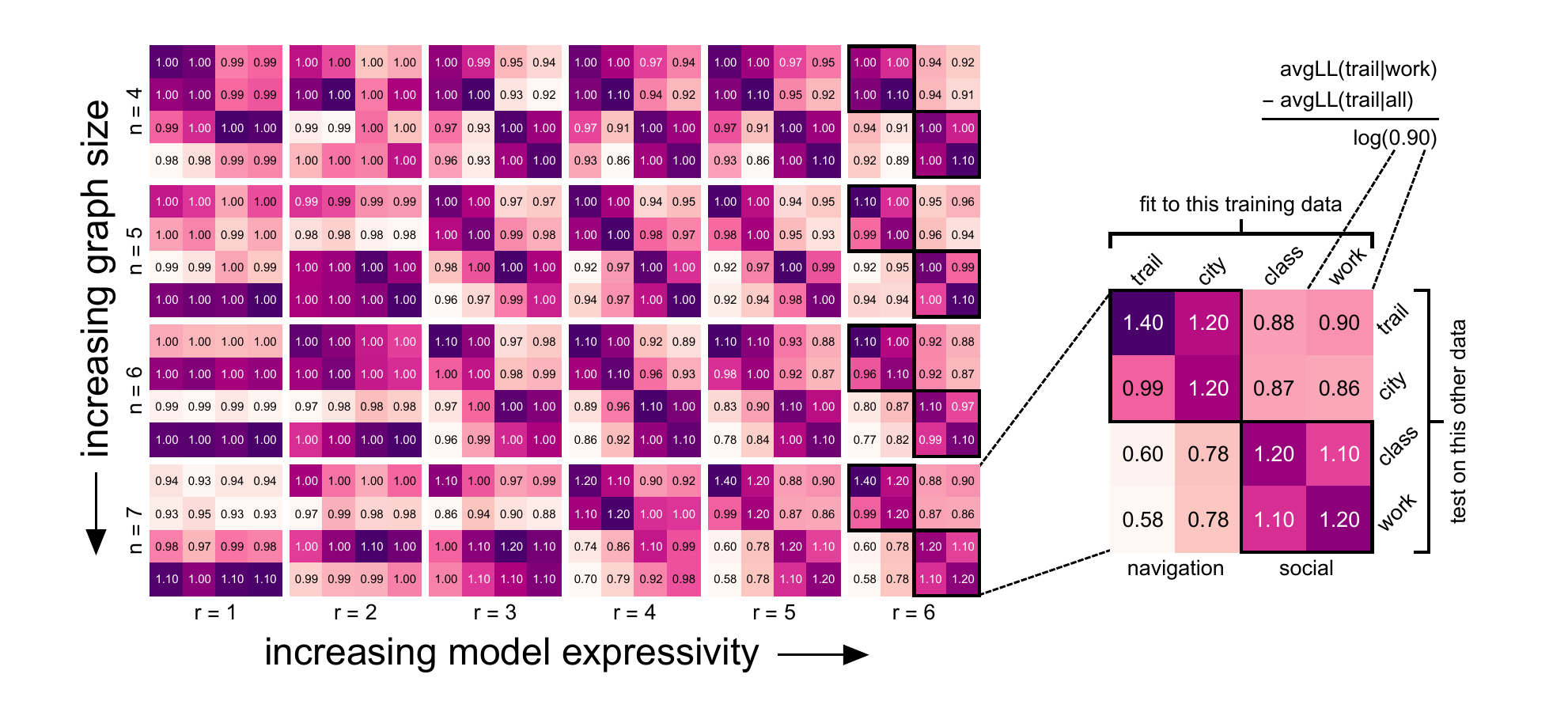}}
\vspace{-0.1in}
\caption{\textbf{Increasing the expressivity of the model for the priors reveals \mbox{domain-specific} traits.}\\ 
The \mbox{$4$-by-$6$} layout divides the results by the number of nodes \smash{\mbox{$4\leq n\leq7$}} (rows) and the model complexity \smash{\mbox{$1\leq r\leq6$}} (columns) of the priors. 
Each of these $24$ options for $n$ and $r$ contains results summarized by a \mbox{$4$-by-$4$} square of numbers. 
For each of these \mbox{$4$-by-$4$} square, we fit $4$ ``specialized'' models 
(using the training data of each cover story separately), 
as well as one ``\mbox{non-specialized}'' model 
(using the combined training data of all $4$ cover stories).  
As seen in the amplified \mbox{$4$-by-$4$} square on the right, 
the $4$ columns denote the cover story of the specialized training data, 
and the $4$ rows denote the cover story of the test data used to evaluate these models. 
The numbers inside these squares are defined in section~\ref{ssec:generalizationresults} and rounded to the nearest $0.01$.  
They can be approximately thought of as an average \mbox{odds-ratio}; 
a value of \mbox{$1\pm\varepsilon$} corresponds to the case when the specialized model 
assigns a probability to an actual response of a participant 
that is about \mbox{$1\pm\varepsilon$} times the probability assigned by the \mbox{non-specialized} model. 
The coloring of the squares is purely for visualization; 
we scale the same range of colors to that square's maximum and minimum values. 
While the less expressive models (to the left) are in fact much more similar than the colors suggest, 
the more expressive models recover priors that are better able to differentiate between domains and (to a lesser degree) specific cover stories. 
}
\label{fig:comparisonwithfitorder}
\end{center}
\end{figure*}

\finalSolve{In figure~\ref{fig:comparisonwithfitorder}, we compare generalization within domain and between domains.} 
In particular, for a given number of nodes $n$ and model expressivity $r$ (equation~\ref{eq:LowDimensionalPrior}) for the priors, 
we randomly partitioned the data from each of the $4$ cover stories into a training set ($80\%$) and a test set ($20\%$). 
Then, for each of the \smash{\mbox{$4\times4=16$}} combinations of cover stories, we fit the (order $r$) model to the training data and measured the average  \mbox{log-likelihood} per round (which we denote by ``{\avgLL}'') of the test data under this model.  
To compare to a meaningful baseline, we subtracted the
{\avgLL} 
of this test data under a ``\mbox{non-specialized}'' model that was fit to the combined training data of all four cover stories.
%
%

\finalSolve{The decimal numbers shown in figure~\ref{fig:comparisonwithfitorder} are the exponential of these differences in {\avgLL}, having ``units'' of a ratio of probabilities.  
A value of $1.00$ corresponds to the specialized model explaining the data equally as well as the \mbox{non-specialized} model, while a value greater than one indicates that the specialized model explains the data better than the \mbox{non-specialized} model (and conversely for a value less than one). 
Figure~\ref{fig:comparisonwithfitorder} shows the average result for $64$ repetitions of this procedure.} 

\ParagraphVertSpace
\textbf{Larger subgraphs reveal differences between domains. }
As a general trend, we find that more complex models recover priors that are better able to differentiate between domains and (to a lesser degree) specific cover stories.  
This is reflected in figure~\ref{fig:comparisonwithfitorder} by the suggestively ``\mbox{block-diagonal}'' appearance of the \mbox{$4$-by-$4$} squares corresponding to more expressive models (\smash{\mbox{$r\gtrsim3$}}). 
\finalSolve{The ``\mbox{horizontal-row}'' appearance of the \mbox{$4$-by-$4$} squares  corresponding to \smash{\mbox{$r\lesssim2$}} suggests that the quality of fit for less complex models is determined primarily by the particular data used for testing.} 
These results are in agreement with our previous findings that larger motifs (and triangles in particular) are needed in order to distinguish between the two domains (figs.~\ref{fig:sparsepriors},~\ref{fig:egalitarian}, and~\ref{fig:socialtriangles}). 

\ParagraphVertSpace
\textbf{A note on planar graphs. }
One aspect worth mentioning is the duality between our two spatial navigation cover stories. 
While both are suggestively planar, their connectivity is of two different flavors. 
The trails in the \trailcolor{park} cover story are rather analogous to \mbox{vectors} (a ``large'' trail implies a large separation between the two nature sites), while the boundaries between neighborhoods in the \citycolor{city} cover story are analogous to \mbox{one-forms} (a ``large'' boundary between neighborhoods implies that they are nearby).  
While in our results, the similarity between the priors for these two navigation cover stories is about the same as the similarity between those for the two social cover stories, it is possible that future investigations involving weighted graphs could reflect this difference.  
\vspace{-10pt}

\section{Possible Sequels}
\label{sec:FutureDirection}
%
\textbf{Other structures. }
In addition to weighted graphs, other extensions are also possible.  
For example, one could investigate priors over bipartite graphs representing people's preferences over a set of items, or priors over directed graphs modeling patterns of citations. 
More generally, any such MCMCP experiment might benefit from our approach of inferring the prior by explicitly fitting the assumed Bayesian model to the aggregated data.  

\ParagraphVertSpace
\textbf{Adaptive sampling. }
As our method explicitly uses the MCMCP assumptions to fit the data, there is no longer a need to collect data in chains. 
In fact, one could use the current fit of the prior to inform an adaptive sampling algorithm to select the {\InfoInWord} presented in each iteration. 
As a simple example, we found that it was helpful to initialize many chains over a large range of edge densities.  
It is entirely possible that a similar ``spreading out'' over other features (like degree heterogeneity and clustering) would likewise be helpful. 

\ParagraphVertSpace
\textbf{Larger graphs. }
Our results suggest that properties of the priors vary with the number of nodes. 
The results presented in the main text consider graphs with at most \mbox{$8$} nodes.  
Unfortunately, in practice, our experiments appeared to lose human engagement for graphs with $10$ or more nodes (see discussion in appendix~\ref{app:scalability} and \ref{sec:EdgeBimodal}, and fig.~\ref{fig:edgedensityallnodes} for analysis of these data). 
Adapting our methods to measure graphs over a range of sizes would be particularly interesting, as
we could compare the resulting priors with actual structure of analogous real networks.
The scaled graph cumulants we used in our analysis are \mbox{well-suited} for such comparisons. 

\ParagraphVertSpace
\textbf{Different cultures. }
It is important to note that if we are to claim that a prior is representative of general characteristics of human cognition, it should be representative of the full diversity of the humans.  In this direction, we have an ongoing collaboration with a linguist that works with the \textit{Yawanaw\'a} and the \textit{Xinane} aboriginal tribes in the Amazon rainforest \cite{baker2020switch}.  
In the domain of navigation, it would be interesting to see if their priors change when the discussion is about paths versus when the discussion is about regions.  
Results from the social domain may also prove interesting; in both tribes, while \mbox{parallel-cousins} (i.e., their parents are \mbox{same-sex} siblings) are forbidden from marriage, marriage between \mbox{cross-cousins} (i.e., their parents are \mbox{opposite-sex} siblings) is considered ideal. 
Case studies such as this could offer insight into the effects of community size and social norms on our priors over social networks. 

\ParagraphVertSpace
\textbf{General message. }
This paper offers a \mbox{case-study} about the power of carefully constructed experiments and clever analysis.  
Just as neural networks benefit from having architecture that reflects the symmetry of the data \cite{villar2023passive}, 
so too does the design and analysis of experiments that use people as substrate.  
The results presented here demonstrate two such examples:  
how the assumed Bayesian structure of the participants' responses can be used to more efficiently use the collected data, 
and how the relabelling symmetry of relational data can be leveraged to design more computationally tractable models for their priors.  
%
%

\vspace{-8pt}
\section*{Acknowledgements}
\vspace{-3pt}
I did this work as part of my PhD thesis at the Princeton Neuroscience Institute (PNI). 
I acknowledge PNI for the financial support during my PhD and the Princeton's Cognitive Science Department for an independent research grant. 
%
The completion of this work is inextricably connected to the discussions and support shared with me by \textit{many} incredible people along the
PhD journey (see my acknowledgments in \citet{bravo2020quantifying}). 
Here, I would like to particularly thank: 
\href{https://scholar.google.com/citations?user=UAwKvEsAAAAJ&hl=en}{Tom Griffiths}; 
\href{https://scholar.google.com/citations?user=tFrElIUAAAAJ&hl=en}{Talmo Pereira}, for his essential role in building such an amazing online game platform; 
and \href{https://scholar.google.com/citations?user=dXZ2pDsAAAAJ&hl=en}{\mbox{Lee~M.~Gunderson}}, whose insights and feedback permeate every bit of this work. 
\nocite{deutsch2011beginning, bartlett1932remembering, montague2007your,lake2017building,doya2007bayesian,rieke1999spikes,thompson2021human,morgan2020experimental, canini2014revealing,lee2017determing,sanborn2018representational,huang2011predictive,wark2007sensory,botvinick2015reinforement, harrison2020gibbs, yamakoshi2022probing,marjieh2022analyzing}

\nocite{rieke1999spikes, manookin2023two,frydman2022efficient, bhui2018decision,
bayes1763anessay,canini2014revealing}
\bibliography{MCMCPBib}
\bibliographystyle{icml2023}

\newpage
\appendix
\onecolumn

\icmltitlerunning{Quantifying Human Priors over Social and Navigation Networks --- \textit{Supplementary Information}}

\newpage
\vspace{12pt}
\section{Experimental Procedure}
\label{app:experimentalprocedure}

In this section, we provide a detailed description of our experiments, and protocols for data collection and cleaning. 

\vspace{12pt}
\subsection{Data Collection}
\label{sec:datacollection}

All participants were recruited online using Amazon Mechanical Turk (AMT) (see e.g. \citet{crowston2012amazon} for a description of the AMT system). 
We only recruited participants who doing our experiment for the first time and had at least $90\%$ of their completed ``HITs'' (i.e., experiments intermediated by the AMT crowdsourcing system) approved. 

The experiments were approved by Princeton University’s Institutional Review Board (IRB) for human subjects, and all participants provided informed consent for the study. 

The experiment lasted $38$ minutes on average and
participants were paid an average wage rate of \(\$11\) per hour.

\vspace{12pt}
\subsection{Experimental Design}
\label{app:experimentaldesign}

We developed a ``gamified'' online experimental platform that smoothly allocates participants to the appropriate experimental chains in real time.

The structure of the experiments was the same for all four cover stories (table~\ref{table:coverstories}). 
In the the two social cover stories:
\vspace{-6pt}
\begin{enumerate}
\renewcommand{\labelenumi}{\textit{\theenumi}}
\renewcommand{\theenumi}{\textcolor{Blue}{\textit{\arabic{enumi}}.}}
\item \classcolor{\textbf{Class}}: participants were asked to infer the \classcolor{friendships} (\textit{relations}) between \classcolor{students} (\textit{nodes}) in a \classcolor{classroom} (\textit{context}). \\\vspace{-14pt}
\item \workcolor{\textbf{Work}}: participants were asked to infer the \workcolor{friendships} (\textit{relations}) between \workcolor{coworkers} (\textit{nodes}) in a \workcolor{workplace} (\textit{context}). \\\vspace{-14pt}
\end{enumerate}
\vspace{-2pt}
And in the two navigation cover stories:
\vspace{-6pt}
\begin{enumerate}
\item \trailcolor{\textbf{Park}}: participants were asked to infer the \trailcolor{trails} (\textit{relations}) between \trailcolor{nature sites} (\textit{nodes}) in a \trailcolor{nature park} (\textit{context}). \\[-14pt]
\item \citycolor{\textbf{City}}: participants were asked to infer the \citycolor{borders} (\textit{relations}) between \citycolor{neighborhoods} (\textit{nodes}) in a \citycolor{city} (\textit{context}). \\[-8pt]
\end{enumerate}
\vspace{-8pt}

Each experiment began with an introduction about the particular cover story. 
It then posed several questions to the participant to ensure their understanding (see appendix~\ref{app:screenshots} for the full text for each of the cover stories).
After that, the task/game started. 
It consisted of a series of ``rounds''.

A round proceeded as follows (see \href{https://www.youtube.com/watch?v=aZNeN293MZs}{\underline{here}} for a demonstration video):
\vspace{-4pt}
\begin{itemize}
\item \text{\textbf{Main interface page}} (see the screenshots in figs.~\ref{fig:ScreenshotSocStudent} and~\ref{fig:ScreenshotNavTrail}): \\
In the center of the screen, there was a visualization of the graph and an interactive interface.\\
Using this interface, the participant could:\\ move the nodes, add edges to the graph, and remove edges from the graph. \\
The nodes were initially positioned in such a way that the nodes did not overlap and the edges were not ambiguous.\footnote{This was achieved using a modified \mbox{spring-electrical} model for graph drawing \cite{hu2011algorithms}, with an additional penalty for edges with the same slope.} 
Connections that were not obscured were already placed in the graph,\\ 
along with a list of the ``unobscured'' relations at the top of the screen. \\
The most important points of the introduction for properly doing the experiment were also recalled in this page. \\ 
Once the participant was satisfied with their modifications,\\they submitted their response by clicking the ``Done!'' button at the bottom right of the screen. 
\\\vspace{-12pt}
\item \text{\textbf{\mbox{Post-round} engagement page}}: \\
To foster engagement, once the participant submitted their response,\\a question (asked in a variety of ways) appeared about which node(s) they thought to be the most/least important. \\ 
The participant was shown the graph they had just submitted, and gave an answer\\by clicking on the node(s) they thought were the most/least important before clicking the ``Submit'' button. \\\vspace{-14pt}
\end{itemize}
\vspace{-2pt}

There were $16$ rounds in total in an experiment, i.e., a participant (potentially)\footnote{See appendix~\ref{app:exclusioncriteria} for the exclusion criteria we used to decide whether to append a response to a chain.} contributed to $16$ \textit{different} chains, thus completing $16$ different graphs.
However, a participant could, of course, quit the experiment at any point.  
In such cases, we still had their data up to that point recorded and we compensated the participant for the work they had completed.
\vspace{4pt}

These $16$ rounds consisted of $2$ rounds for each number of nodes \mbox{$n \in \{4,5,6,7,8,10,12,15\}$} with a varying number of relations shown $\numShown$ (i.e., the number of relations that were not obscured out of the total number of pairwise relations \mbox{$\numRelations \in \{6,10,15,21,28,45,66,105\}$}).

In particular, for each cover story,
participants were randomly assigned to one of the following six options for the precise sequence of rounds:
\vspace{-6pt}
\begin{enumerate}
\renewcommand{\labelenumi}{\textit{\theenumi}}
\renewcommand{\theenumi}{\textcolor{Blue}{\textit{\arabic{enumi}}.}}
\item $(n,\numShown)$: $(4,2)$, $(4,4)$, $(5,3)$, $(5,7)$, $(6,5)$, $(6,7)$, $(7,6)$, $(7,9)$, $(8,7)$, $(8,12)$, $(10,8)$, $(10,19)$, $(12,8)$, $(12,28)$, $(15,20)$, $(15,40)$; \\\vspace{-12pt}
\item $(n,\numShown)$: $(4,4)$, $(4,2)$, $(5,7)$, $(5,3)$, $(6,7)$, $(6,5)$, $(7,9)$, $(7,6)$, $(8,12)$, $(8,7)$, $(10,19)$, $(10,8)$, $(12,28)$, $(12,8)$, $(15,40)$, $(15,20)$; \\\vspace{-12pt}
\item $(n,\numShown)$: $(4,5)$, $(5,5)$, $(6,7)$, $(7,9)$, $(8,12)$, $(10,19)$, $(12,28)$, $(15,10)$, $(4,3)$, $(5,1)$, $(6,5)$, $(7,6)$, $(8,7)$, $(10,8)$, $(12,15)$, $(15,10)$; \\\vspace{-12pt}
\item $(n,\numShown)$: $(4,3)$, $(5,1)$, $(6,5)$, $(7,6)$, $(8,7)$, $(10,8)$, $(12,15)$, $(15,10)$, $(4,5)$, $(5,5)$, $(6,7)$, $(7,9)$, $(8,12)$, $(10,19)$, $(12,28)$, $(15,10)$; \\\vspace{-12pt}
\item $(n,\numShown)$: $(4,3)$, $(5,9)$, $(6,7)$, $(7,9)$, $(8,12)$, $(10,19)$, $(12,15)$, $(15,40)$, $(15,10)$, $(12,8)$, $(10,8)$, $(8,7)$, $(7,6)$, $(6,5)$, $(5,5)$, $(4,1)$; or \\\vspace{-12pt}
\item $(n,\numShown)$: $(4,1)$, $(5,5)$, $(6,5)$, $(7,6)$, $(8,7)$, $(10,8)$, $(12,8)$, $(15,10)$, $(15,40)$, $(12,15)$, $(10,19)$, $(8,12)$, $(7,9)$, $(6,7)$, $(5,9)$, $(4,3)$. \\\vspace{-14pt}
\end{enumerate}

We added participants to the chains until they contained $12$ participants.  
When needed, we initialized a new chain with a new random graph, sampled in a way that ensured that the initial graphs covered a large range of edge densities. 
Only responses that passed our exclusion criteria (described in appendix~\ref{app:exclusioncriteria}) were appended to the chain.

For each of the cover stories, 
we ran the experiments at least until we obtained 
two chains of length $12$ for all the $16$ $(n,\numShown)$ pairs in each of the six different sequences.\footnote{For some of the chains over graphs with $\gtrsim10$ nodes, it took quite a few rounds of participants to obtain a response.}
While there are no results for graphs with $10$, $12$, and $15$ nodes in the main text (for reasons discussed in appendix~\ref{app:scalability}), 
we were able to model the density of connections for these larger graphs (as described in appendix~\ref{sec:EdgeBimodal}). 
\newpage

\vskip 0.3in 
\begin{figure}[H] 
\vspace{100pt}
	\begin{center}
	\includegraphics[trim={0cm 0cm 0.03cm 0cm},clip,width=1\columnwidth]{Figures/ScreenshotMainpageClass.pdf}
    \caption{\textbf{Screenshot of the main interface page of our experiment for the social \classcolor{class} cover story.
    }}
      \label{fig:ScreenshotSocStudent}
	 \end{center}	
\end{figure}
\newpage

\vskip 0.3in 
\begin{figure}[H] 
\vspace{100pt}
	\begin{center}
    \includegraphics[trim={0cm 0cm 0.cm 0cm},clip,width=1\columnwidth]{Figures/ScreenshotMainpageTrail.pdf}
   \caption{\textbf{Screenshot of the main interface page of our experiment for the navigation \trailcolor{park} cover story.
   }}
  \label{fig:ScreenshotNavTrail}
	 \end{center}	
\end{figure}
\newpage

\vspace{12pt}
\subsection{Design Considerations}
\label{app:appexperimentalconsiderations}

We performed a variety of pilot experiments (totaling more than $300$ participants), which provided valuable insight into how to make these experiments engaging and intuitive.  

For example, the first version had no visual interface for manipulating the graphs, and the participants had to remember the relations while responding to a series of yes/no questions.  
As our experiments are not particularly concerned about \mbox{short-term} memory, 
removing this unnecessary and cognitively taxing obstacle proved very helpful.  
We also added an extra question after each graph to make it more engaging, 
as well as many embellishments to the cover stories.

These and other improvements were incorporated into the final experiments, resulting in remarkably positive feedback in the \mbox{post-experiment} questionnaire, as well as several MTurk workers sending personal emails about how they found the experiments engaging.

These pilot experiments also allowed us to devise quantitative heuristics to clean the data (appendix~\ref{app:exclusioncriteria}), thus (hopefully) only including ``genuine'' responses to the chains. 

\vspace{12pt}
\subsection{Exclusion Criteria}
\label{app:exclusioncriteria}

Data from behavioral experiments with humans, particularly when collected online, 
can be ``contaminated'' by participants that are not sufficiently engaged with the experiment. 
Thus, we implemented a systematic method for excluding such data from the chains.
We also use this method for rewarding participants that clearly gave thoughtful deliberation to our experiments. 

Our exclusion heuristics were judiciously chosen after observing the distribution of participants’ responses to our pilot experiments. 
Specifically, we exclude all rounds that met any of the following criteria:
\vspace{-6pt}
\begin{enumerate}
\item \textbf{\textit{Answered too quickly}}:\\if the participant took less than $3$ seconds per shown relation to submit their response. \\\vspace{-12pt}
\item \textbf{\textit{Not enough interaction}}:\\if the participant moved fewer than \mbox{$\left \lceil \frac{n}{4} \right \rceil - 1$} nodes. \\\vspace{-12pt}
\item \textbf{\textit{Changed too little}}:\\if \mbox{$\numObs > 5$} and \mbox{$f_{\text{add}} \times n < 1$},\\where $\numObs$ is the number of relations obscured, $f_{\text{add}}$ is the fraction of obscured relations that the participant choose to be edges in their response, and $n$ is the number of nodes.\\\vspace{-12pt}
\item \textbf{\textit{Not enough practice}}:\\if the participant had fewer than $4$ valid rounds.\\\vspace{-10pt}
\end{enumerate}

The total number of rounds and the total number of participants for each cover story before and after the exclusion criteria are displayed in table~\ref{table:numdatapoints}. 
\begin{table}[H]
\caption{\textbf{Amount of data before and after applying our exclusion criteria.}\\ 
After applying the exclusion criteria, we used approximately \mbox{$90 \%$} of the total number of data points (i.e., rounds).  
}
\vskip 0.1in 
{\renewcommand{\arraystretch}{1.5}
\begin{tabular}{c|c|c}
    \tableFontTop{cover story}                     &  $\numsymbol$ \tableFontTop{ participants:}  \tableFontTopBefAf{after out of total (\% excluded)}    &   $\numsymbol$ \tableFontTop{ rounds:} \tableFontTopBefAf{after out of total (\% excluded)}  \\
    \hline
    \cline{1-3}
      \tableFontStoryNameBefAft{\classcolor{class}}   &   
                    \tablenumdatapropfont{443}{362}  \tablenumdataperc{18}  &         \tablenumdatapropfont{5340}{4795} \tablenumdataperc{10}\\

    \hline
          \tableFontStoryNameBefAft{\workcolor{work}}   &   
                      \tablenumdatapropfont{342}{269}   \tablenumdataperc{21}  &   \tablenumdatapropfont{4317}{3675} \tablenumdataperc{15}\\
             \hline
          \tableFontStoryNameBefAft{\trailcolor{park}}   &   
                      \tablenumdatapropfont{359}{299}  \tablenumdataperc{17}    &   \tablenumdatapropfont{4163}{3823} \tablenumdataperc{8}\\

    \hline
             \tableFontStoryNameBefAft{\citycolor{city}}   &   
                      \tablenumdatapropfont{347}{289}  \tablenumdataperc{17}    &   \tablenumdatapropfont{4013}{3569} \tablenumdataperc{11}  \\

\end{tabular}
}
\label{table:numdatapoints}
\end{table}

\newpage

\vspace{12pt}
\subsection{Detailed Instructions}
\label{app:screenshots}

In this section, we provide the entire instruction text, page by page, for each of the four cover stories.
The instructions were broken in several pages, and participants could navigate to the next page or the previous page.

\vspace{12pt}
\subsubsection{General Format}
\label{sss:instructiongenral}

All experiments started with the same \textbf{welcome page}:

\begin{quote}
Thank you and welcome to our experiment!
\spacequote

Next, we will show you a few \textbf{instructions}.\\
Please read them carefully,\\as you will have to \textbf{correctly answer}\\a few questions before moving on to the game!
\end{quote}

After the instructions,
we asked participants three multiple choice questions to verify that they understood the task. 
Each question appeared on a single page, and 
participants were only allowed to move to the next page once they had answered the question correctly. 
If they answered correctly, they would simply see a message displaying:  
\begin{quote}
    \textbf{Correct!}
\end{quote}

If they answered incorrectly, 
they would see a ``wrong answer message page'' with a summary of the cover story they were participating in. 
This message was the same for the three questions (but, of course, different for each cover story). 

For all cover stories, 
the correct answer for the first question was option $2$, for the second question was option $1$, and for the third question was option $2$.

We now provide the text specific to each cover story.

\vspace{12pt}
\subsubsection{Cover Story: \classcolor{\textbf{Class}}}
\label{sec:detailsclassstory}

\begin{center}
     \classcolor{\textbf{\textsc{Instructions}}}
\end{center}

\spacepageinstruction
\paragraph{\classcolor{Page $\bm{1}$:}} 

\begin{quote}
We are studying how gossip spreads in schools.
\spacequote

In a variety of different classes,\\we \textbf{recorded the friendships} between pairs of students.
\spacequote

We are testing how well people \textbf{intuit} these friendship networks\\based on \textbf{partial information}.
\end{quote}

\newpage
\spacepageinstruction
\paragraph{\classcolor{Page $\bm{2}$:}} 

\begin{quote}
    In each class:
\spacequote

\textbf{Some} pairs of students are \textbf{friends}.\\
So, gossip can be directly transferred between these two students\\ without needing to pass through another student.
\spacequote

\textbf{Other} pairs of students are \textbf{not friends}.\\
So, for gossip to be transferred from one to the other,\\it has to pass through at least one other student.
\spacequote

You will play the following game:
\spacequote

\begin{center}
    \textbf{We tell you whether some pairs of students are friends or not.}
    
\textbf{Your goal is to reconstruct the rest of their friendship network.}
    %
\end{center}
\spacequote

You \textbf{win points} by matching the ``shape/structure'' of the unknown relations!
\end{quote}

\spacepageinstruction
\paragraph{\classcolor{Page $\bm{3}$:}} 

\begin{quote}
For each round of the game,\\
we \textbf{randomly select} students from the \textbf{same class},\\
and display \textbf{some} of their relations at the top of the screen.
\spacequote

For example:
\begin{itemize}
\item ``Hassen and Hernandez are friends''

\item ``Miller and Fleming are NOT friends''
\end{itemize}

\spacequote
\textbf{But the list is incomplete!}
\spacequote

You need to use your \textbf{social intuition} and \textbf{reasoning} to\\decide whether the other pairs are friends or not.
\spacequote

You will do this by \textbf{drawing} the rest of their \textbf{friendship} network\\using our graphical interface.
\end{quote}

\spacepageinstruction
\paragraph{\classcolor{Page $\bm{4}$:}}

\begin{quote}
       \textbf{How to draw the friendship network:}
    \spacequote

       \begin{sc}
   \textcolor{gray}{[Here we had a quick video with a demo of the interface]}
   \end{sc}

    \spacequote
    \begin{itemize}
        \item To \textbf{change the location} of a student, \textbf{click and drag} their name.
        \spacequote
        
        \item To \textbf{connect} two students, first \textbf{click on one, then on the other}.\\
A \textbf{line} will appear between them, indicating that they are \textbf{friends}.
\spacequote

\item To \textbf{disconnect} two students,\\simply \textbf{click on the line} that connects them.\\
The line will disappear, indicating that they are \textbf{not friends}.
\spacequote
    \end{itemize}
\spacequote
     
     Notes:
     \spacequote
     
     If there is \textbf{no line} between two students,\\it means you think they are \textbf{not friends}.\\
\textit{Even if in your drawing they look very close to each other!}\\
So, if you think two students are \textbf{friends},\\always make sure to connect them with a \textbf{line}.
   \spacequote
   
   To make your job easier,\\
we have already connected the friendship pairs from the list for you.\\
And if you attempt to connect a non-friendship pair from the list,\\
we indicate the error with a red ``X''.
\spacequote

We start the students at \textbf{random positions},\\
so make sure to move the students around,\\
as this will help you visualize the network.
\end{quote}

\spacepageinstruction
\paragraph{\classcolor{Page $\bm{5}$:}} 

\begin{quote}
Some important information:
\spacequote

\begin{itemize}
\item You will play this game for \textbf{several rounds},\\
each time with a \textbf{different class}.
\spacequote

\item In each round, the students are \textbf{randomly} selected from the \textbf{same} class.
\spacequote

\item The friendships were recorded from \textbf{actual} classrooms,\\
so to protect the students' identities, we use \textbf{fictitious names}.\\
Thus, the names do \textbf{not provide any information}\\and you should not use them to guide your answers.
\spacequote

\item To motivate you to do your best,\\you will be \textbf{paid} according to your \textbf{performance},
which is determined by\\how well your drawings match the \textbf{actual} friendship networks.
\spacequote

\item Precisely, we will keep a \textbf{score} for each round:\\
You \textbf{win points} for correctly inferring if\\the pairs of students \textbf{not presented} in the list are friends or not.\\
You \textbf{lose points} if your drawing does \textbf{not respect}\\the relations given in the list, which you know for sure are correct.\\
The closer you match the \textbf{actual} friendship networks,\\the \textbf{larger} your bonus will be.\\
We will give your total \textbf{score} and the resulting performance bonus\\only at the \textbf{end} of the experiment.
\spacequote

\item We will give you a chance to take a \textbf{break} at the \textbf{end} of each round.\\Please attempt to solve each round \textbf{uninterrupted}.
   \end{itemize} 
   \spacequote
\end{quote}

\newpage
\spacequote
\begin{center}
          \classcolor{\textbf{\textsc{Questions after instructions}}}
\end{center}
\spacequote

\spacepageinstruction
\paragraph{\classcolor{Wrong answer message:}} 
\begin{quote}
    \textbf{Sorry, but...}
    \spacequote

We recorded the \textbf{friendships} between pairs of \textbf{students},\\
and we are testing how well people \textbf{intuit} these friendship networks\\based on \textbf{partial information}.
\spacequote

In particular, for each round,\\
we randomly select some students from the same class,\\
and tell you whether \textbf{some} pairs of students are friends or not.
\spacequote

Your \textbf{goal} is to reconstruct the rest of their friendship network.
\end{quote}

\spacepageinstruction
\paragraph{\classcolor{Page for question $\bm{1}$:}}

\begin{quote}
Before we move on,\\please answer a few quick questions to make sure you understand the game.\\
Feel free to use the \textbf{Previous} button if you need to review the instructions.
\spacequote

Here's an easy one to get started:
 \spacequote

 \textbf{What are we asking you to draw?}
 \spacequote
 \begin{enumerate}
\renewcommand{\theenumi}{\formatInstructionNumberQuestions{\arabic{enumi}}}
     \item Power grid networks.
      \spacequote
      
     \item Student friendship networks.
      \spacequote
 \end{enumerate}
 
\end{quote}

\spacepageinstruction
\paragraph{\classcolor{Page for question $\bm{2}$:}} 

\begin{quote}
\textbf{What do you know about the relations between students?}
\spacequote

\begin{enumerate}
\renewcommand{\theenumi}{\formatInstructionNumberQuestions{\arabic{enumi}}}
    \item Some pairs of students are \textbf{friends},\\meaning gossip can transfer directly between them.\\
Other pairs are \textbf{not friends},\\so gossip must pass through \textbf{at least} one other student to get between them.
\spacequote

 \item Some pairs of students are in the \textbf{same class},\\meaning they know each other.\\
Other pairs of students are in \textbf{different classes},\\which means they likely \textbf{do not} know each other.
\spacequote

 \item Some pairs of students are in the \textbf{same school},\\meaning they possibly know each other.\\
Other pairs of individuals are in \textbf{different schools},\\which means they do not know each other.
\spacequote
\end{enumerate}

\end{quote}

\spacepageinstruction
\paragraph{\classcolor{Page for question $\bm{3}$:}} 

\begin{quote}
\textbf{What is your goal, and what are its main challenges?}
\spacequote

\begin{enumerate}
\renewcommand{\theenumi}{\formatInstructionNumberQuestions{\arabic{enumi}}}
    \item Your goal is to discover which classes are \textbf{dysfunctional},\\
and therefore more likely to support bullying and bad behavior.\\
The main challenge is that you do not know who these students are\\or the schools they come from.
 \spacequote

\item 
 Your goal is to reconstruct the friendship network\\of randomly selected students.\\
The main challenge is that we only tell you\\whether \textbf{some} pairs of students are friends or not.
 \spacequote
\end{enumerate}
\end{quote}

\spacepageinstruction
\paragraph{\classcolor{Final instruction page:}} 

\begin{quote}
  Awesome job! You are now ready to reconstruct your first friendship network!  
\end{quote}
\vspace{12pt}


\subsubsection{Cover Story: \workcolor{\textbf{Work}}}
\label{sec:detailsworkstory}

\begin{center}
    \workcolor{\textbf{\textsc{Instructions}}}
\end{center}

\spacepageinstruction
\paragraph{\workcolor{Page $\bm{1}$:}} 

\begin{quote}
We are studying how gossip spreads in workplaces.
\spacequote

In a variety of different workplaces,\\we \textbf{recorded the friendships} between pairs of coworkers.
\spacequote

We are testing how well people \textbf{intuit} these friendship networks\\based on \textbf{partial information}.
\end{quote}

\spacepageinstruction
\paragraph{\workcolor{Page $\bm{2}$:}} 

\begin{quote}
    In each workplace:
\spacequote

\textbf{Some} pairs of coworkers are \textbf{friends}.\\
So, gossip can be directly transferred between these two coworkers\\ without needing to pass through another coworker.
\spacequote

\textbf{Other} pairs of coworkers are \textbf{not friends}.\\
So, for gossip to be transferred from one to the other,\\ 
it has to pass through at least one other coworker.
\spacequote

You will play the following game:
\spacequote

\begin{center}
    \textbf{We tell you whether some pairs of coworkers are friends or not.}
    
 \textbf{Your goal is to reconstruct the rest of their friendship network.}
    %
\end{center}
\spacequote

You \textbf{win points} by matching the ``shape/structure'' of the unknown relations!
\end{quote}

\spacepageinstruction
\paragraph{\workcolor{Page $\bm{3}$:}} 

\begin{quote}
For each round of the game,\\
we \textbf{randomly select} coworkers from a \textbf{single workplace},\\
and display \textbf{some} of their relations at the top of the screen.
\spacequote

For example:
\begin{itemize}
\item ``Hassen and Hernandez are friends''

\item ``Miller and Fleming are NOT friends''
\end{itemize}

\spacequote
\textbf{But the list is incomplete!}
\spacequote

You need to use your \textbf{social intuition} and \textbf{reasoning} to\\decide whether the other pairs are friends or not.
\spacequote

You will do this by \textbf{drawing} the rest of their \textbf{friendship} network\\using our graphical interface.
\end{quote}

\spacepageinstruction
\paragraph{\workcolor{Page $\bm{4}$:}}

\begin{quote}
       \textbf{How to draw the friendship network:}
    \spacequote

   \begin{sc}
   \textcolor{gray}{[Here we had a quick video with a demo of the interface]}
   \end{sc}
   
    \spacequote
    \begin{itemize}
        \item To \textbf{change the location} of a person, \textbf{click and drag} their name.
        \spacequote
        
        \item To \textbf{connect} two coworkers, first \textbf{click on one, then on the other}.\\
A \textbf{line} will appear between them, indicating that they are \textbf{friends}.
\spacequote

\item To \textbf{disconnect} two coworkers,\\simply \textbf{click on the line} that connects them.\\
The line will disappear, indicating that they are \textbf{not friends}.
\spacequote
    \end{itemize}
\spacequote
     
     Notes:
     \spacequote
     
     If there is \textbf{no line} between two coworkers,\\it means you think they are \textbf{not friends}.\\
\textit{Even if in your drawing they look very close to each other!}\\
So, if you think two coworkers are \textbf{friends},\\always make sure to connect them with a \textbf{line}.
   \spacequote
   
   To make your job easier,\\
we have already connected the friendship pairs from the list for you.\\
And if you attempt to connect a non-friendship pair from the list,\\
we indicate the error with a red ``X''.
\spacequote

We start the coworkers at \textbf{random positions},\\
so make sure to move them around,\\
as this will help you visualize the network.
\end{quote}

\spacepageinstruction
\paragraph{\workcolor{Page $\bm{5}$:}} 

\begin{quote}
Some important information:
\spacequote

\begin{itemize}
\item You will play this game for \textbf{several rounds},\\each time with a \textbf{different workplace}.
\spacequote

\item In each round, the coworkers are \textbf{randomly} selected from a \textbf{single} workplace.
\spacequote

\item The friendships were recorded from \textbf{actual} workplaces,\\
so to protect their identities, we use \textbf{fictitious names}.\\
Thus, the names do \textbf{not provide any information}\\and you should not use them to guide your answers.
\spacequote

\item To motivate you to do your best,\\you will be \textbf{paid} according to your \textbf{performance},
which is determined by\\how well your drawings match the \textbf{actual} friendship networks.
\spacequote

\item Precisely, we will keep a \textbf{score} for each round:\\
You \textbf{win points} for correctly inferring if\\the pairs of coworkers \textbf{not presented} in the list are friends or not.\\
You \textbf{lose points} if your drawing does \textbf{not respect}\\the relations given in the list, which you know for sure are correct.\\
The closer you match the \textbf{actual} friendship networks,\\the \textbf{larger} your bonus will be.\\
We will give your total \textbf{score} and the resulting performance bonus\\only at the \textbf{end} of the experiment.
\spacequote

\item We will give you a chance to take a \textbf{break} at the \textbf{end} of each round.\\Please attempt to solve each round \textbf{uninterrupted}.
   \end{itemize} 
   \spacequote
\end{quote}

\spacequote
\begin{center}
        \workcolor{\textbf{\textsc{Questions after instructions}}}
\end{center}
\spacequote

\spacepageinstruction
\paragraph{\workcolor{Wrong answer message:}} 
\begin{quote}
    \textbf{Sorry, but...}
    \spacequote

We recorded the \textbf{friendships} between pairs of \textbf{coworkers},\\
and we are testing how well people \textbf{intuit} these friendship networks\\based on \textbf{partial information}.
\spacequote

In particular, for each round,\\
we randomly select some people from a single workplace,\\
and tell you whether \textbf{some} pairs of coworkers are friends or not.
\spacequote

Your \textbf{goal} is to reconstruct the rest of their friendship network.
\end{quote}

\newpage
\spacepageinstruction
\paragraph{\workcolor{Page for question $\bm{1}$:}}

\begin{quote}
Before we move on,\\please answer a few quick questions to make sure you understand the game.\\
Feel free to use the \textbf{Previous} button if you need to review the instructions.
\spacequote

Here's an easy one to get started:
 \spacequote
 
 \textbf{What are we asking you to draw?}
 \spacequote
 \begin{enumerate}
 \renewcommand{\theenumi}{\formatInstructionNumberQuestions{\arabic{enumi}}}
     \item Power grid networks.
      \spacequote
      
     \item Coworker friendship networks.
      \spacequote
 \end{enumerate}
 
\end{quote}

\spacepageinstruction
\paragraph{\workcolor{Page for question $\bm{2}$:}} 

\begin{quote}
\textbf{What do you know about the relations between these people?}
\spacequote

\begin{enumerate}
\renewcommand{\theenumi}{\formatInstructionNumberQuestions{\arabic{enumi}}}
    \item Some pairs of coworkers are \textbf{friends},\\meaning gossip can transfer directly between them.\\
Other pairs are \textbf{not friends},\\so gossip must pass through \textbf{at least} one other coworker to get between them.
\spacequote

 \item Some pairs of people are in the \textbf{same workplace},\\meaning they know each other.\\
Other pairs of people are in \textbf{different workplaces},\\which means they likely \textbf{do not} know each other.
\spacequote

 \item Some pairs of people are in the \textbf{same company},\\meaning they possibly know each other.\\
Other pairs of people are in \textbf{different companies},\\which means they do not know each other.
\spacequote
\end{enumerate}

\end{quote}

\spacepageinstruction
\paragraph{\workcolor{Page for question $\bm{3}$:}} 

\begin{quote}
\textbf{What is your goal, and what are its main challenges?}
\spacequote

\begin{enumerate}
\renewcommand{\theenumi}{\formatInstructionNumberQuestions{\arabic{enumi}}}
    \item Your goal is to discover which workplaces are \textbf{dysfunctional},\\
and therefore more likely to have low employee satisfaction.\\
The main challenge is that you do not know who these people are\\or where they work.
 \spacequote
 
\item 
 Your goal is to reconstruct the friendship network\\of randomly selected coworkers.\\
The main challenge is that we only tell you\\whether \textbf{some} pairs of coworkers are friends or not.
 \spacequote
\end{enumerate}
\end{quote}

\spacepageinstruction
\paragraph{\workcolor{Final instruction page:}} 

\begin{quote}
  Awesome job! You are now ready to reconstruct your first friendship network!  
\end{quote}
\vspace{12pt}

\subsubsection{Cover Story: \trailcolor{\textbf{Park}}}
\label{sec:detailstrailstory}

\begin{center}
    \trailcolor{\textbf{\textsc{Instructions}}}
\end{center}

\spacepageinstruction
\paragraph{\trailcolor{Page $\bm{1}$:}} 

\begin{quote}
We \textbf{recorded the trail maps}  of several nature parks,\\and you will be visiting a different park in each round of this game.
\spacequote

As part of planning for the trip, you studied the trail map\\and created an exciting list of places to go.
\spacequote

But as you arrive there, you realize you \textbf{forgot the map} at home\\(and due to bugdet cuts, there is not a single map there)!
\end{quote}

\spacepageinstruction
\paragraph{\trailcolor{Page $\bm{2}$:}} 

\begin{quote}
    In each nature park:
\spacequote

\textbf{Some pairs} of nature sites have a \textbf{direct trail} connecting them.\\
So you can go from one to the other \textbf{directly} from one to the other,\\ without passing through any other nature site.
\spacequote

\textbf{Other} pairs of nature sites do \textbf{not} have a direct trail connecting them.\\
So, to go from one to the other,\\you must pass through at least one other nature site.
\spacequote

You will play the following game:
\spacequote

\begin{center}
    \textbf{We tell you whether some pairs of nature sites have a direct trail connecting them or not.}
    
\textbf{
Your goal is to reconstruct the rest of the trail map.}
    %
\end{center}
\spacequote

You \textbf{win points} by matching the ``shape/structure'' of the unknown trails!
\end{quote}

\spacepageinstruction
\paragraph{\trailcolor{Page $\bm{3}$:}} 

\begin{quote}
For each round of the game,\\
we \textbf{randomly select} nature sites from a \textbf{single nature park},\\
and display \textbf{some} of their relations at the top of the screen.
\spacequote

For example:
\begin{itemize}
\item ``There is a direct trail connecting Hassen and Hernandez''

\item ``There is NO direct trail connecting Miller and Fleming''
\end{itemize}

\spacequote
\textbf{But the list is incomplete!}
\spacequote

You need to use your \textbf{navigation skills} and \textbf{spatial reasoning} to\\decide whether the other pairs of nature sites have direct trails connecting them or not.
\spacequote

You will do this by \textbf{drawing} the rest of the \textbf{trail map} using our graphical interface.
\end{quote}

\spacepageinstruction
\paragraph{\trailcolor{Page $\bm{4}$:}}

\begin{quote}
      \textbf{How to draw the trail map:}
    \spacequote

       \begin{sc}
   \textcolor{gray}{[Here we had a quick video with a demo of the interface]}
   \end{sc}

    \spacequote
    \begin{itemize}
        \item To \textbf{change the location} of a nature site, \textbf{click and drag} it.
        \spacequote
        
        \item To \textbf{connect} two nature sites, first \textbf{click on one, then on the other}.\\
A \textbf{line} will appear between them, indicating that there is a \textbf{direct trail} connecting the two.
\spacequote

\item To \textbf{disconnect} two nature sites, simply \textbf{click on the line} that connects them.\\
The line will disappear, indicating that there is \textbf{no direct trail} connecting them.
\spacequote
    \end{itemize}
\spacequote
     
     Notes:
     \spacequote
     
     If there is \textbf{no line} between two nature sites,\\it means you think that there is \textbf{no direct trail} connecting them.\\
\textit{Even if in your drawing they look very close to each other!}\\
So, if you think that there is a \textbf{direct trail} connecting two nature sites,\\always make sure to connect them with a \textbf{line}.
   \spacequote
   
To make your job easier,\\
we have already connected for you the pairs from the list that have a direct trail connecting them.\\
And if you attempt to connect a pair from the list that has no direct trail connecting them,\\
we indicate the error with a red ``X''.

\spacequote

We start the nature sites at \textbf{random positions},\\
so make sure to move them around,\\
as this will help you visualize and more accurately reconstruct the map.
\end{quote}

\spacepageinstruction
\paragraph{\trailcolor{Page $\bm{5}$:}} 

\begin{quote}
Some important information:
\spacequote

\begin{itemize}
\item You will play this game for \textbf{several rounds},\\each time with a \textbf{different nature park}.
\spacequote

\item In each round, the nature sites are \textbf{randomly} selected from a \textbf{single} nature park.
\spacequote

\item These are trails from \textbf{actual} nature parks,\\
but we use \textbf{fictitious names} for the nature sites,\\so that the game cannot be solved by a simple google search!\\
Thus, the names \textbf{do not provide any information}\\ and you should not use them to guide your answers.
\spacequote

\item To motivate you to do your best,\\you will be \textbf{paid} according to your \textbf{performance},
which is determined by\\how well your drawings match the \textbf{actual} trail maps.
\spacequote

\item Precisely, we will keep a \textbf{score} for each round:\\
You \textbf{win points} for correctly inferring if\\the pairs of nature sites \textbf{not presented} in the list have a direct trail connecting them or not.\\
You \textbf{lose points} if your drawing does \textbf{not respect}\\the relations given in the list, which you know for sure are correct.\\
The closer you match the \textbf{actual} trail maps,\\the \textbf{larger} your bonus will be.\\
We will give your total \textbf{score} and the resulting performance bonus\\only at the \textbf{end} of the experiment.
\spacequote

\item We will give you a chance to take a \textbf{break} at the \textbf{end} of each round.\\Please attempt to solve each round \textbf{uninterrupted}.
   \end{itemize} 
   \spacequote
\end{quote}

\spacequote
\begin{center}
     \trailcolor{\textbf{\textsc{Questions after instructions}}}
\end{center}
\spacequote

\spacepageinstruction
\paragraph{\trailcolor{Wrong answer message:}} 
\begin{quote}
    \textbf{Sorry, but...}
    \spacequote

We recorded the \textbf{trail map} of several nature parks,\\
and you must navigate them using only \textbf{partial information}.
\spacequote

In particular, for each round,\\
we randomly select some trails from a single park,\\
and tell you whether \textbf{some} pairs of sites have a direct trail connecting them or not.
\spacequote

Your \textbf{goal} is to reconstruct the rest of the trail map.
\end{quote}

\spacepageinstruction
\paragraph{\trailcolor{Page for question $\bm{1}$:}}

\begin{quote}
Before we move on,\\please answer a few quick questions to make sure you understand the game.\\
Feel free to use the \textbf{Previous} button if you need to review the instructions.
\spacequote

Here's an easy one to get started:
 \spacequote
 
 \textbf{What are we asking you to draw?}
 \spacequote
 \begin{enumerate}
 \renewcommand{\theenumi}{\formatInstructionNumberQuestions{\arabic{enumi}}}
     \item Maps of train stations.
     \spacequote
     
     \item Maps of nature parks. 
     \spacequote
 \end{enumerate}
 
\end{quote}

\spacepageinstruction
\paragraph{\trailcolor{Page for question $\bm{2}$:}} 

\begin{quote}
\textbf{What do you know about the nature sites?}
\spacequote

\begin{enumerate}
\renewcommand{\theenumi}{\formatInstructionNumberQuestions{\arabic{enumi}}}
    \item Some pairs have a \textbf{direct trail} connecting them,\\
so you can go from one to the other without passing through any other nature site.\\
Other pairs do \textbf{not} have a direct trail connecting them,\\
so you must pass through \textbf{at least} one other site to go from one to the other. 
\spacequote

 \item  Some pairs have a \textbf{direct trail} connecting them,\\
so they are \textbf{close} to each other.\\
Other pairs do \textbf{not} have a direct trail connecting them,\\
so these two nature sites are \textbf{far apart}.
\spacequote

 \item  Some pairs have a \textbf{direct trail} connecting them,\\
so they are in the \textbf{same} nature park.\\
Other pairs do \textbf{not} have a trail connecting them,\\
so they are in a \textbf{different} nature park.
\spacequote

\end{enumerate}

\end{quote}

\spacepageinstruction
\paragraph{\trailcolor{Page for question $\bm{3}$:}} 

\begin{quote}
\textbf{What is your goal, and what are its main challenges?}
\spacequote

\begin{enumerate}
\renewcommand{\theenumi}{\formatInstructionNumberQuestions{\arabic{enumi}}}
    \item Your goal is to discover the \textbf{shortest} path that visits \textbf{all} the nature sites.
\\
The main challenge is that you do not know\\how far apart the nature sites are from each other.
\spacequote

\item 
Your goal is to draw the trail map\\of randomly selected nature sites from a single nature park.\\
The main challenge is that we only tell you whether\\ \textbf{some} pairs of them are connected by a direct trail or not. 
\spacequote
\end{enumerate}
\end{quote}


\spacepageinstruction
\paragraph{\trailcolor{Final instruction page:}} 

\begin{quote}
  Awesome job!  You are now ready to reconstruct your first trail map!
\end{quote}
\vspace{12pt}

\subsubsection{Cover Story: \citycolor{\textbf{City}}}
\label{sec:detailscitystory}

\begin{center}
    \citycolor{\textbf{\textsc{Instructions}}}
\end{center}

\spacepageinstruction
\paragraph{\citycolor{Page $\bm{1}$:}} 

\begin{quote}
We \textbf{recorded the neighborhood map} of several cities,\\and you will be visiting a different city in each round of this game.
\spacequote

As part of the trip planning, you studied the city map\\and created an exciting list of places to go.
\spacequote

But as you arrive in the city, you realize you \textbf{forgot your map} at home!\\
Fortunately, you partially remember the layout,\\and immediately begin filling in the rest.
\end{quote}

\spacepageinstruction
\paragraph{\citycolor{Page $\bm{2}$:}} 

\begin{quote}
    In each city:
\spacequote

\textbf{Some} pairs of neighborhoods \textbf{share a border}.\\
So crossing it allows you to go \textbf{directly} from one to the other,\\
without passing through any other neighborhood.
\spacequote

\textbf{Other} pairs of neighborhoods do \textbf{not} share a border.\\
So, to go from one to the other,\\you must pass through at least one other neighborhood.
\spacequote

You will play the following game:
\end{quote}

\begin{center}
\textbf{
We tell you whether some pairs of neighborhoods share a border or not. 
}

\textbf{
Your goal is to reconstruct the rest of the neighborhood map.}
\end{center}

\begin{quote}
You \textbf{win points} by matching the ``shape/structure'' of the unknown borders!
\end{quote}

\spacepageinstruction
\paragraph{\citycolor{Page $\bm{3}$:}} 

\begin{quote}
For each round of the game,\\
we \textbf{randomly select} neighborhoods from a \textbf{single city},\\
and display \textbf{some} of their relations at the top of the screen.
\spacequote

For example:
\begin{itemize}
\item ``Hassen and Hernandez share a border''

\item ``Miller and Fleming do NOT share a border''
\end{itemize}

\spacequote
\textbf{But the list is incomplete!}
\spacequote

You need to use your \textbf{navigation skills} and \textbf{spatial reasoning} to\\decide whether the other pairs of neighborhoods share a border or not.
\spacequote

You will do this by \textbf{drawing} the rest of the \textbf{neighborhood map}\\using our graphical interface.
\end{quote}

\spacepageinstruction
\paragraph{\citycolor{Page $\bm{4}$:}}

\begin{quote}
    \textbf{How to draw the neighboorhood map:}
    \spacequote

       \begin{sc}
   \textcolor{gray}{[Here we had a quick video with a demo of the interface]}
   \end{sc}

    \spacequote
    \begin{itemize}
        \item To \textbf{change the location} of a neighborhood, \textbf{click and drag} it.
        \spacequote
        
        \item To \textbf{connect} two neighborhoods, first \textbf{click on one, then on the other}.\\
A \textbf{line} will appear between them, indicating that they \textbf{share a border}.
\spacequote

\item To \textbf{disconnect} two neighborhoods,\\simply \textbf{click on the line} that connects them.\\
The line will disappear, indicating that they do \textbf{not share a border}.
\spacequote
    \end{itemize}
\spacequote
     
     Notes:
     \spacequote
     
     If there is \textbf{no line} between two neighborhoods,\\it means you think that they do \textbf{not share a border}.\\
\textit{Even if in your drawing they look very close to each other!}\\
So, if you think two neighborhoods \textbf{share a border},\\always make sure to connect them with a \textbf{line}.
   \spacequote
   
To make your job easier,\\
we have already connected for you the pairs from the list that share a border.\\
And if you attempt to connect a pair from the list that does not share a border,\\
we indicate the error with a red ``X''.
\spacequote

We start the neighborhoods at \textbf{random positions},\\
so make sure to move them around,\\
as this will help you visualize and more accurately reconstruct the map.
\end{quote}

\spacepageinstruction
\paragraph{\citycolor{Page $\bm{5}$:}} 

\begin{quote}
Some important information:
\spacequote

\begin{itemize}
\item You will play this game for \textbf{several rounds},\\each time with a \textbf{different city}.
\spacequote

\item In each round, the neighborhoods are \textbf{randomly} selected from a \textbf{single} city.
\spacequote

\item These are neighborhoods from \textbf{actual} cities,\\
but we use \textbf{fictitious names} for the neighborhoods,\\so that the game cannot be solved by a simple google search!\\
Thus, the names \textbf{do not provide any information}\\ and you should not use them to guide your answers.
\spacequote

\item To motivate you to do your best,\\you will be \textbf{paid} according to your \textbf{performance},
which is determined by\\how well your drawings match the \textbf{actual} neighborhood maps.
\spacequote

\item Precisely, we will keep a \textbf{score} for each round:\\
You \textbf{win points} for correctly inferring if\\the pairs of neighborhoods \textbf{not presented} in the list share a border or not.\\
You \textbf{lose points} if your drawing does \textbf{not respect}\\the relations given in the list, which you know for sure are correct.\\
The closer you match the \textbf{actual} neighborhood maps,\\the \textbf{larger} your bonus will be.\\
We will give your total \textbf{score} and the resulting performance bonus\\only at the \textbf{end} of the experiment.
\spacequote

\item We will give you a chance to take a \textbf{break} at the \textbf{end} of each round.\\Please attempt to solve each round \textbf{uninterrupted}.
   \end{itemize} 
   \spacequote
\end{quote}

\spacequote
\begin{center}
     \citycolor{\textbf{\textsc{Questions after instructions}}}
\end{center}
\spacequote

\spacepageinstruction
\paragraph{\citycolor{Wrong answer message:}} 
 \begin{quote}
    \textbf{Sorry, but...}
    \spacequote

We recorded the \textbf{neighborhood map} of several cities,\\
and you must navigate them using only \textbf{partial information}.
\spacequote

In particular, for each round,\\
we randomly select some neighborhoods from a single city,\\
and tell you whether \textbf{some} pairs of neighborhood share a border or not.
\spacequote

Your \textbf{goal} is to reconstruct the rest of the neighborhood map.
 \end{quote}

\spacepageinstruction
\paragraph{\citycolor{Page for question $\bm{1}$:}}

\begin{quote}
Before we move on,\\please answer a few quick questions to make sure you understand the game.\\
Feel free to use the \textbf{Previous} button if you need to review the instructions.
\spacequote

Here's an easy one to get started:
 \spacequote
 
 \textbf{What are we asking you to draw?}
 \spacequote
 \begin{enumerate}
 \renewcommand{\theenumi}{\formatInstructionNumberQuestions{\arabic{enumi}}}
     \item Subway maps of stations.
      \spacequote
      
     \item City maps of neighborhoods.
      \spacequote
 \end{enumerate}
 
 \end{quote}

\spacepageinstruction
\paragraph{\citycolor{Page for question $\bm{2}$:}} 

\begin{quote}
\textbf{What do you know about the neighborhoods?}
\spacequote

\begin{enumerate}
\renewcommand{\theenumi}{\formatInstructionNumberQuestions{\arabic{enumi}}}
    \item Some pairs \textbf{share a border},\\so crossing it allows you to go \textbf{directly} from one to the other.\\
Other pairs do \textbf{not} share a border,\\so you must pass through \textbf{at least} one other neighborhood to go from one to the other.
\spacequote

 \item Some pairs \textbf{share a border},\\so they are \textbf{close} to each other.\\
Other pairs do \textbf{not} share a border,\\so these two neighborhoods are \textbf{far apart}.
\spacequote

 \item Some pairs \textbf{share a border},\\so they are \textbf{similar} to each other.\\
Other pairs do \textbf{not} share a border,\\so these two neighborhoods are very \textbf{different}.
\spacequote
\end{enumerate}

\end{quote}


\spacepageinstruction
\paragraph{\citycolor{Page for question $\bm{3}$:}} 

\begin{quote}
\textbf{What is your goal, and what are its main challenges?}
\spacequote

\begin{enumerate}
\renewcommand{\theenumi}{\formatInstructionNumberQuestions{\arabic{enumi}}}
    \item Your goal is to discover the \textbf{shortest} path that visits \textbf{all} the neighborhoods.
\\
The main challenge is that you do not know\\how far apart the neighborhoods are from each other.
 \spacequote

\item 
 Your goal is to draw the neighborhood map\\of randomly selected neighborhoods from a single city.\\
The main challenge is that we only tell you whether\\ \textbf{some} pairs of them share a border or not.
 \spacequote
\end{enumerate}
\end{quote}

\spacepageinstruction
\paragraph{\citycolor{Final instruction page:}} 

\begin{quote}
  Awesome job!  You are now ready to reconstruct your first neighborhood map!
\end{quote}

\newpage
\vspace{12pt}
\section{Modeling the Experimental Data}
\label{app:modelling}

In this section, we describe in detail how we fit the data from our experiments and selected the models for their priors. 

\vspace{12pt}
\subsection{Model Fitting}
\label{app:fittingprocedure}

For each cover story and each number of nodes, 
we fit the MCMCP Bayesian model to the participants' aggregated data using a natural parameterization for the priors (equation~\ref{eq:LowDimensionalPriorAppendix}).
The only free parameters of the model are those parameterizing the prior. 

While it is technically possible to fit a model to the data from each individual chain separately,
we chose to aggregate the data of multiple chains. 
Besides increasing the statistical power (by increasing the number of data points), this also helps  
obtain data over a larger space of graphs and remove potential effects of initial conditions. 
There are two reasons for this: 
\vspace{-5pt}
\begin{enumerate}
\renewcommand{\labelenumi}{\textit{\theenumi}}
\renewcommand{\theenumi}{\textcolor{Blue}{\textit{\arabic{enumi}}.}}
    \item The initial graphs were sampled in a way that enforced a large range of edge density. Thus, by aggregating data from multiple chains, we have a large variety of initial conditions. \\\vspace{-12pt}
    \item The number of relations obscured varied between the chains, 
    which might in practice influence the space of graphs that the participants considered (although when we split the data by fraction of relations obscured, the inferred priors did not appear to have any significant trend).  \\\vspace{-12pt}
\end{enumerate}
\vspace{-0pt}
 
We modeled the priors using a hierarchical family of maximum entropy distributions over simple graphs\footnote{Recall that by ``simple graphs'' we mean: unweighted and undirected graphs that do not have \mbox{self-loops} or multiple parallel edges.} with $n$ nodes (section~\ref{sec:maximumentropypriors}).  
For completeness we recall this model here:
\vspace{-0pt}
\begin{equation}
    \Prior\big(G\big)\,\propto\,\ERrhohalf\big(G\big)\,\times\,\exp\Bigg\{\sum_{g:E(g)\leq r} \beta_g^{ } \mu_g^{ } \big(G\big) \Bigg\}
    \label{eq:LowDimensionalPriorAppendix}
\end{equation}

where the constrained statistics are the injective homomorphism densities {\mugText} (footnote~\ref{foot:injhom}) of all subgraphs $g$ with \mbox{$\leq r$} edges.\footnote{That is, all subgraphs with $r$ edges, including disconnected subgraphs with no isolated nodes.}
Thus, these distributions describe a nested family of models for networks indexed by the parameter $r$. 
We call $r$ the ``order'' of this model, it corresponds to the expressivity/complexity of the model.

When modeling the priors to these models, we constrained these subgraph densities {\mugText} to match their measured value in the data by fitting the Lagrangian parameters {\betagText} associated with them. 

In particular, we maximized the \mbox{log-likelihood} of participants' data under this model, 
which is given by: 
\vspace{-0pt}
\begin{align}
\log\kern-1pt\Big(\mathcal{L}\big(\Data|\vect{\beta}\kern2pt\big)\kern-2pt\Big) = \mathlarger{\sum_t} \Bigg[\log\kern-1pt\bigg( \ERrhohalf\big(G_t^{ }\kern1pt|\kern1pt\PartialGraph_t^{ }\big) \TimesGap \!\exp\!\Big\{\vect{\beta}\cdot\vect{\mu}\big(G_t^{ }\big)\!\Big\} \bigg) - \log\kern-1pt\bigg(
\sum\limits_{G' \in \GraphSet_{\kern-1pt n}^{ }} \ERrhohalf\big(G'\kern1pt|\kern1pt\PartialGraph_t^{ }\big) \TimesGap \!\exp\!\Big\{\vect{\beta}\cdot\vect{\mu}\big(G'\big)\!\Big\} \bigg)\Bigg]
\label{eq:loglikelihooddataappendix}
\vspace{-12pt}
\end{align}
where 
\vspace{-7pt}
 \begin{itemize}
 \item \mbox{$\ERrhohalf\big(G'\kern1pt|\kern1pt\PartialGraph_t^{ }\big)$} indicates a restriction of the \mbox{fully-random} distribution {\ERrhohalff} to only the relations that were obscured at round $t$, thereby restricting to the graphs that were possible responses on round $t$; \\\vspace{-15pt}
 \item $\mathcal{G}_n$ is the set of simple graphs with $n$ nodes; and \\\vspace{-15pt}
 \item  the sum in equation~\ref{eq:LowDimensionalPriorAppendix} has been summarized as the dot product between the parameters {\betagText} and subgraph densities {\mugText}. 
 \end{itemize}
\newpage
To simplify notation, let \smash{\mbox{$\mathcal{L\!L} := \log\big(\mathcal{L}(\Data|\vect{\beta})\big)$}} denote equation~\ref{eq:loglikelihooddataappendix}. 
We maximized $\mathcal{L\!L}$ using Newton's method. 

The entries of the gradient $\vect{\nabla}\mathcal{L\!L}$ are given by: 
\begin{align}
\frac{\partial\mathcal{L\!L}}{\partial\beta_i^{ }} = \mathlarger{\sum_t} \left[ \mu_{g(i)}^{ }\MuShrinkGap\big(G_t^{ }\big) - \frac{\mathlarger{\sum_{G'\in \GraphSet_{\kern-1pt n}^{ }}}\mu_{g(i)}^{ }\MuShrinkGap\big(G'\big) \TimesGap \ERrhohalf\big(G'\kern1pt|\kern1pt\PartialGraph_t^{ }\big) \TimesGap \!\exp\!\Big\{\vect{\beta}\cdot\vect{\mu}\big(G'\big)\!\Big\}}{\mathlarger{\sum_{G'\in \GraphSet_{\kern-1pt n}^{ }}} \ERrhohalf\big(G'\kern1pt|\kern1pt\PartialGraph_t^{ }\big) \TimesGap \!\exp\!\Big\{\vect{\beta}\cdot\vect{\mu}\big(G'\big)\!\Big\}} \right] 
\label{eq:gradloglikelihooddata}
\end{align}

where $\mu_{g(i)}^{ }$ is the subgraph density associated with the parameter $\beta_i^{ }$.

And the entries of the matrix of second derivatives $\vect{\nabla}\vect{\nabla}\mathcal{L\!L}$ are given by: 
\begin{align}
&\frac{\partial^2\mathcal{L\!L}}{\partial\beta_i^{ }\partial\beta_j^{ }} = \nonumber\\
&\mathlarger{\sum_t} \left[  \frac{ \Bigg(\mathlarger{\sum_{G'\in \GraphSet_{\kern-1pt n}^{ }}}\mu_{g(i)}^{ }\MuShrinkGap\big(G'\big) \TimesGap \ERrhohalf\big(G'\kern1pt|\kern1pt\PartialGraph_t^{ }\big) \TimesGap \!\exp\!\Big\{\vect{\beta}\cdot\vect{\mu}\big(G'\big)\!\Big\}\Bigg) \Bigg(\mathlarger{\sum_{G'\in \GraphSet_{\kern-1pt n}^{ }}}\mu_{g(j)}^{ }\MuShrinkGap\big(G'\big) \TimesGap \ERrhohalf\big(G'\kern1pt|\kern1pt\PartialGraph_t^{ }\big) \TimesGap \!\exp\!\Big\{\vect{\beta}\cdot\vect{\mu}\big(G'\big)\!\Big\}\Bigg)}{\Bigg( \mathlarger{\sum_{G'\in \GraphSet_{\kern-1pt n}^{ }}} \ERrhohalf\big(G'\kern1pt|\kern1pt\PartialGraph_t^{ }\big) \TimesGap \!\exp\!\Big\{\vect{\beta}\cdot\vect{\mu}\big(G'\big)\!\Big\} \Bigg)^{\kern-2pt 2}} \right. \nonumber\\
&\qquad\qquad \left. -  \quad \frac{ \mathlarger{\sum_{G'\in \GraphSet_{\kern-1pt n}^{ }}}\mu_{g(i)}^{ }\MuShrinkGap\big(G'\big) \TimesGap \mu_{g(j)}^{ }\MuShrinkGap\big(G'\big) \TimesGap \ERrhohalf\big(G'\kern1pt|\kern1pt\PartialGraph_t^{ }\big) \TimesGap \!\exp\!\Big\{\vect{\beta}\cdot\vect{\mu}\big(G'\big)\!\Big\} }{\mathlarger{\sum_{G'\in \GraphSet_{\kern-1pt n}^{ }}}\ERrhohalf\big(G'\kern1pt|\kern1pt\PartialGraph_t^{ }\big) \TimesGap\!\exp\!\Big\{\vect{\beta}\cdot\vect{\mu}\big(G'\big)\!\Big\}}
\vphantom{\frac{ 
            \Bigg(\mathlarger{\sum_{G'\in \GraphSet_{\kern-1pt n}^{ }}}\mu_{g(i)}^{ }\big(G'\big) \ERrhohalf\big(G'\kern1pt|\kern1pt\PartialGraph_t^{ }\big) \exp\!\Big\{\vect{\beta}\cdot\vect{\mu}\big(G'\big)\Big\}\Bigg) \Bigg(\mathlarger{\sum_{G'\in \GraphSet_{\kern-1pt n}^{ }}}\mu_{g(j)}^{ }(G') \ERrhohalf\big(G'\kern1pt|\kern1pt\PartialGraph_t^{ }\big) \exp\!\Big\{\vect{\beta}\cdot\vect{\mu}\big(G'\big)\Big\}\Bigg)}{\Bigg( \mathlarger{\sum_{G'\in \GraphSet_{\kern-1pt n}^{ }}} \ERrhohalf\big(G'\kern1pt|\kern1pt\PartialGraph_t^{ }\big) \exp\!\Big\{\vect{\beta}\cdot\vect{\mu}\big(G'\big)\Big\} \Bigg)^{\kern-2pt 2}}}  
\right]. 
\label{eq:hessianloglikelihooddata}
\end{align}

We then repeated the Newton iteration,  \mbox{$\vect{\beta} \leftarrow \vect{\beta} - \Big( \vect{\nabla} \vect{\nabla} \mathcal{L\!L} \Big)^{\kern-2pt-1} \kern-1pt\cdot\kern1pt \Big( \vect{\nabla}\mathcal{L\!L} \Big )$},  until machine precision.  

\vspace{12pt}
\subsubsection{Maximum Entropy Priors over the Number of Connections Only}
\label{app:secedgesonlymethods}

In figure~\ref{fig:edgedensityallnodes} we show results for priors over the distribution of number of edges only. 
Our procedure for obtaining these priors was essentially the same as above. 
The only difference is that 
instead of using maximum entropy priors over simple graphs with $n$ nodes,
we used maximum entropy priors over the set of binary sequences of length \smash{\mbox{$\binom{n}{2}$}}.
The constrained statistics for these maximum entropy models are the moments of these sequences, i.e., the expectations of powers of the density of ones. 
\vspace{5pt}


\vspace{12pt}
\subsection{Scalability}
\label{app:scalability}
When fitting priors over graphs with $7$ nodes or less, we enumerated all the possibilities  explicitly (i.e., all the valid $G'$ for \mbox{$\ERrhohalf\big(G'\kern1pt|\kern1pt\InfoInOne_t^{ }\big)$} in equation~\ref{eq:loglikelihooddataappendix}).

For larger graphs, to handle the combinatorial explosion inherent with an increasing number of nodes (see table~\ref{table:numgraphs} for a visualization of the scale), we employed a method of subsampling graphs from {\ERrhohalff} with appropriate weights.  
This allowed us to fit distributions over graphs with $8$ nodes. 

For example, in some cases, we obscured $21$ relations (out of the $28$), resulting in $2^{21}$ possible ways to complete the graph, thereby necessitating such a method.

\newpage
\begin{table}[H]
\caption{\textbf{A combinatorial explosion.\\} 
For the number of nodes displayed in the \tableFontTop{nodes} column,
the \tableFontTop{relations} column displays the number of pairwise relations (edges and non-edges) for simple graphs with that number of nodes (i.e., \smash{$\binom{n}{2}$}), 
the \tableFontTop{unique graphs} column displays the number of nonisomorphic simple graphs with that number of nodes,
and the \tableFontTop{unique representations} column displays the number of simple graphs with that number of \textit{labelled} nodes (or, equivalently, the number of ordered binary sequences of length \smash{$\binom{n}{2}$}). 
}
\vskip 0.1in 
\begin{center}
{\renewcommand{\arraystretch}{1.2}
\begin{tabular}{l|l|l|l}
    \tableFontTop{nodes}                     &   \tableFontTop{relations}    &  \tableFontTop{unique graphs}      &   \tableFontTop{unique representations}\\
    \hline
    \cline{1-4}
      \tableFontEntriesNum{3}   &   
                      \tableFontEntriesNum{3}  &   \tableFontEntriesNum{4}     &   \tableFontEntriesNum{8}\\
      \tableFontEntriesNum{4}   &   
                      \tableFontEntriesNum{6}  &   \tableFontEntriesNum{11}     &   \tableFontEntriesNum{64}\\
      \tableFontEntriesNum{5}   &   
                      \tableFontEntriesNum{10}  &   \tableFontEntriesNum{34}     &   \tableFontEntriesNum{1024}\\
      \tableFontEntriesNum{6}   &   
                      \tableFontEntriesNum{15}  &   \tableFontEntriesNum{156}     &   \tableFontEntriesNum{32768}\\
      \tableFontEntriesNum{7}   &   
                      \tableFontEntriesNum{21}  &   \tableFontEntriesNum{1044}     &   \tableFontEntriesNum{2097152}\\
      \tableFontEntriesNum{8}   &   
                      \tableFontEntriesNum{28}  &   \tableFontEntriesNum{12346}     &   \tableFontEntriesNum{268435456}\\
      \tableFontEntriesNum{9}   &   
                      \tableFontEntriesNum{36}  &   \tableFontEntriesNum{274668}     &   \tableFontEntriesNum{68719476736}\\
      \tableFontEntriesNum{10}   &   
                      \tableFontEntriesNum{45}  &   \tableFontEntriesNum{12005168}     &   \tableFontEntriesNum{35184372088832}\\
      \tableFontEntriesNum{11}   &   
                      \tableFontEntriesNum{55}  &   \tableFontEntriesNum{1018997864}     &   \tableFontEntriesNum{3602879701896396}\\
      \tableFontEntriesNum{12}   &   
                      \tableFontEntriesNum{66}  &   \tableFontEntriesNum{165091172592}     &   \tableFontEntriesNum{73786976294838206464}\\
      \tableFontEntriesNum{13}   &   
                      \tableFontEntriesNum{78}  &   \tableFontEntriesNum{50502031367952}     &   \tableFontEntriesNum{302231454903657293676544}\\
      \tableFontEntriesNum{14}   &   
                      \tableFontEntriesNum{91}  &   \tableFontEntriesNum{29054155657235488}     &   \tableFontEntriesNum{2475880078570760549798248448}\\
                      \tableFontEntriesNum{15}   &   
                      \tableFontEntriesNum{105}  &   \tableFontEntriesNum{31426485969804308768}     &   \tableFontEntriesNum{40564819207303340847894502572032}\\
\end{tabular}
}
\label{table:numgraphs}
\end{center}
\end{table}

We decided to present results for priors over graphs with $10$ or more nodes in the appendix essentially for three reasons: 
\vspace{-5pt}
\begin{enumerate}
\renewcommand{\labelenumi}{\textit{\theenumi}}
\renewcommand{\theenumi}{\textcolor{Blue}{\textit{\arabic{enumi}}.}}
    \item Despite our best efforts in the fitting process, several of the distributions for graphs with $10$ nodes or more\\appeared to become concentrated on the complete graph.\\\vspace{-14pt} 
    \item We have fewer valid data for these nodes (see appendix~\ref{app:exclusioncriteria} for the exclusion criteria),\\but a space that is superexponentially larger. \\\vspace{-14pt}
    \item Some participants indicated in their \mbox{post-questionnaire} that they had difficulties with these rounds (see appendix~\ref{sec:EdgeBimodal}). \\\vspace{-14pt}
\end{enumerate}
 
Still, in certain cases, we managed to obtain priors that appear reasonable. 
So for fun, see \href{https://www.youtube.com/watch?v=siVnYAl9I1Q}{\underline{here}} for an animation of a simulation of a Markov chain using a realistic prior over graphs with $12$ nodes, inferred using data from the social \classcolor{class} cover story (\classcolor{friendships} between \classcolor{students} in a \classcolor{classroom}).

\vspace{12pt}
\subsection{Model Selection and Robustness}
\label{app:modelselection}

For the results in figures~\ref{fig:sparsepriors},~\ref{fig:egalitarian}, and~\ref{fig:socialtriangles},
for each number of nodes and each cover story, 
we selected the order $r$ of the prior by \mbox{cross-validation} using a $80\%$ training set, $20\%$ test set split, and $64$ repetitions of the process.
For all of fit priors, we find that \mbox{higher-order} fits (\mbox{$r = 4$, $5$, or $6$}) were selected.

\paragraph{(W/st)rong assumptions, yet meaningful results.} 
We performed a variety of sanity checks when fitting our models and analyzing the resulting priors. 
For example, we used a number of different splits of the data (including for figure~\ref{fig:comparisonwithfitorder}) and ensured that all results we present in this paper were consistently reproduced. 

\newpage
\vspace{12pt}
\section{Extending to Priors over More Nodes.} 
\label{sec:EdgeBimodal}
Perhaps frustratingly, figures~\ref{fig:sparsepriors}, \ref{fig:egalitarian} and~\ref{fig:socialtriangles} in the main text (section~\ref{sec:ExperimentalResults}) end at graphs with eight nodes.  
While we also collected data for graphs with a larger number of nodes (i.e.,  $10$, $12$, and $15$, see appendix~\ref{app:experimentaldesign}), 
obtaining meaningful results for these graphs presents two major challenges (see appendix~\ref{app:scalability}).
As the number of unique graphs increases\footnote{E.g., there are over \mbox{$12\cdot10^6$} different graphs with $10$ nodes.} 
fitting the model exactly becomes computationally infeasible.  
Additionally, in practice, the attention and engagement of the participants appears to notably decrease when presented with such a large number of constraints and questions. 

\begin{figure}[H] 
\vskip 0.1in
	\begin{center}
 \centerline{\includegraphics[width=1.00\columnwidth]{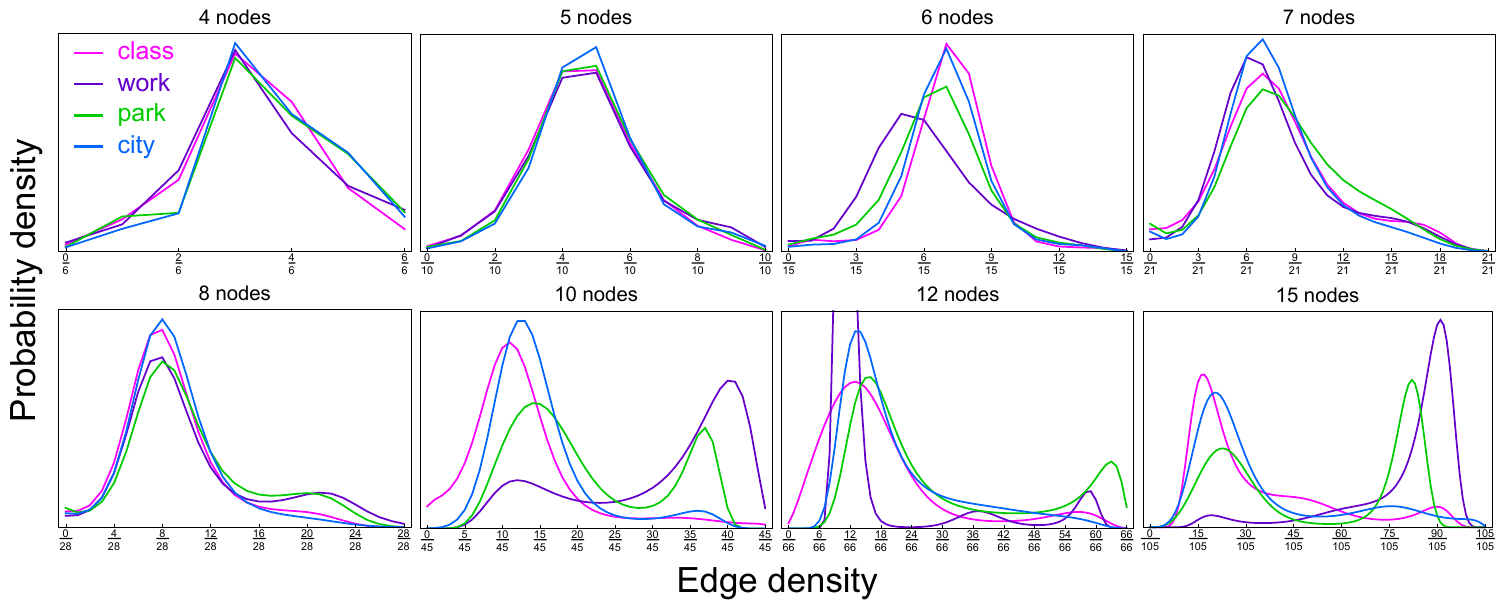}} 
  \vspace{-0.05in}
	\caption{\textbf{Appearance of bimodality in priors over edge density.}\\
 The curves correspond to the inferred priors over the density of connections \textit{only} (i.e., no graphical structure) for a given cover story (indicated by the color of the curve) and a fixed number of nodes (indicated by the title of the subfigure). 
 Notice the general trend from unimodal priors with decreasing mean for smaller graphs, with a second peak of larger edge density appearing for larger graphs.  }
 \label{fig:edgedensityallnodes}
 \end{center}
 \vskip -0.1in	    
\end{figure}

To overcome the computational impasse of larger graphs, 
we consider restricting attention to \textit{only} the edge density (appendix~\ref{app:secedgesonlymethods}).  
Specifically, we model the prior in terms of distributions over the number of edges, 
reducing the domain of the priors to \mbox{${\binom{n}{2}} +1$} (i.e., all graphs with the same number of edges are considered to be the same by the model).  
For the statistics constrained by the maximum entropy parameterization, we used the first six moments of the empirical edge density.  
These results are shown in figure~\ref{fig:edgedensityallnodes}.

\paragraph{Appearance of bimodality in priors over edge density.}
For graphs with \mbox{$n\lesssim8$} nodes, 
these priors consistently have a single peak around an average edge density that decreases with increasing number of nodes (echoing our results in figure~\ref{fig:sparsepriors} in the main text).  
However, for graphs with \mbox{$n\gtrsim10$} nodes, 
these priors become notably bimodal.  

While it is tempting to draw conclusions about this ``regime change'', 
it is worth remembering that human engagement for these larger graphs is still questionable.  
Sparse graphs are easier to remember as they admit a natural compression. 
This compression may equally well be applied to their complements (i.e., the \mbox{nearly-complete} graphs) 
by remembering the ``\mbox{non-edges}''.  
Below \mbox{$n\lesssim8$} nodes, the sparse peak and its dense complement overlap considerably.  
As the number of nodes increases, the dense peak is increasingly \mbox{well-separated} from its sparse complement.  
Thus, it is possible that these results are more a product of the participant's desire to complete the task, 
instead of revealing a profound change in the way we represent graphs with more nodes than we have fingers.  

\newpage
\vspace{12pt}
\section{Markov Chain Monte Carlo with People}
\label{app:mcmcpappendix}

In this section, we first describe the assumptions of the MCMCP model in detail (section~\ref{app:mcmcpassumptions}). 
We then investigate the number of iterations needed for an MCMCP chain to converge sufficiently close to the prior as a function of relevant parameters (section~\ref{app:convergence}).

\vspace{12pt}
\subsection{Assumptions}
\label{app:mcmcpassumptions}

For the sake of completeness, 
we now describe the MCMCP model (section~\ref{sec:MCMCPmodel} and algorithm~\ref{alg:GenericMCMCPExperiment}) and the corresponding assumptions required for the conclusion that the stationary distribution is equal to the prior.  
This discussion can be found in most introduction to probability books covering discrete Markov chains (e.g., \citet{modica2012first}). 

First, let us recall the notation from section~\ref{sec:MCMCPmodel}.  

Let \M{\InfoIn} denote the space of all combinations of {\InfoInWordP} that participants might be given in a MCMCP experimental chain.
Let \M{\InfoOut} denote the space of all {\InfoOutWordP} that participants might consider when giving their responses. 
For simplicity, we consider both $\InfoIn$ and $\InfoOut$ to be discrete and finite (with cardinality $|\InfoIn|$ and $|\InfoOut|$), and denote the space of probability distributions over them as \M{P(\InfoIn)} and \M{P(\InfoOut)}. 

For a given chain in our experiments, 
\mbox{\M{\InfoIn=\PartialGraphSet_{n,\numObs}^{ }}}, the set of all partial graphs with $n$ nodes and $\numObs$ of the $\binom{n}{2}$ pairwise relations obscured.  
The specific partial graph {\PartialGraphText} shown to the participant in the $t^{\text{th}}$ iteration/round is denoted as \mbox{$\PartialGraph_t^{ } \in \PartialGraphSet$}. 
Similarly, \mbox{\M{\InfoOut=\GraphSet_{n}^{ }}}, the set of simple graphs with $n$
nodes. 
The specific simple graph {\GraphText} resulting from the response of the participant in the $t^{\text{th}}$ iteration/round is denoted as \mbox{$\Graph_t^{ } \in \GraphSet$}.

Each round, 
the experimentalist uses the {\InfoOutWord} of the previous participant (i.e., their response) to generate noisy/partial {\InfoInWord} to give to the next participant.  
Let \mbox{ \M{\ExpMapping:\InfoOut\rightarrow\InfoIn}} denote this probabilistic map, 
with associated with probability distribution \mbox{\M{p(\InfoInOne|\InfoOutOne)}}. 

In this setup, the amount of {\InfoInWord} transmitted at each iteration is fixed.\footnote{If the amount of {\InfoInWord} transmitted increases over time, then \mbox{self-sustained} learning can occur (as opposed to convergence to participants' shared prior).  See \citet{chazelle2016self, chazelle2019iterated} for a mathematical analysis of this case.} For example, for our experiments, the number of relations obscured is always the same in a given chain. 

Let \mbox{ \M{\ParticipantMapping:\InfoIn\rightarrow\InfoOut} } be the probabilistic map induced by participants responses, with associated probability distribution \mbox{\M{p(\InfoOutOne|\InfoInOne})}. 
Participants are assumed to be \textit{identical Bayesian agents}, sharing the same prior beliefs and knowledge about the experiment. 

The ``Bayesian'' part of the assumptions refers to the fact that, 
when presented with {\InfoInWordP} \M{\InfoInOne\in\InfoIn}, participants are assumed to respond by \textit{sampling} a hypothesis from their posterior distribution $p(\InfoOutOne|\InfoInOne)$ \\\vspace{-2pt}
\begin{equation}
  p(\InfoOutOne|\InfoInOne) = \frac{p(\InfoInOne|\InfoOutOne)\Prior(\InfoOutOne)}{\sum_{\InfoOutOne \in \InfoOut}p(\InfoInOne|\InfoOutOne)\Prior(\InfoOutOne)}. \label{Eq:Bayes}  
\end{equation}

The ``identical agents'' part of the assumptions implies two assumptions: 
\vspace{-10pt}
\begin{enumerate}
    \item Participants have the \textit{same shared prior} over the {\InfoOutWordP}, \mbox{$\Prior(\InfoOut)$}. \\\vspace{-18pt}
        \item Participants know the \textit{correct likelihood function}  
used to generate the {\InfoInWordP} they observe from a {\InfoOutWordP}, and they use it to compute $p(\InfoInOne|\InfoOutOne)$. \label{commonlikelihoodassumption} \\\vspace{-18pt}
\end{enumerate}
Assumption 
\textit{\textcolor{Blue}{2}} means that the participants are assumed to know the probabilistic map \mbox{ \M{\ExpMapping:\InfoOut\rightarrow\InfoIn}} used by the experimentalist to generate {\InfoInWord} from a {\InfoOutWord}. 
In our experiments, this simply means that they believe that the partial graphs are generated by randomly erasing a fraction of the relations of some underlying graph. 
This is clearly articulated during our experiments (see appendix~\ref{app:screenshots}).

The transition matrix $\TransitionMatrix$ induced by the composed mapping \smash{\mbox{\M{\ParticipantMapping(\ExpMapping(\cdot))\!:P(\InfoOut)\!\rightarrow\!P(\InfoOut)}}} is a \mbox{time-homogeneous} 
Markov chain over the discrete space of {\InfoOutWordPCap}. 

Such a Markov chain converges to a unique stationary distribution if (and only if) it is \textit{ergodic}.  This requires that: 
\vspace{-6pt}
\begin{enumerate}
    \item the chain is irreducible (i.e., any state/hypothesis can be reached from any other state/hypothesis with a \mbox{non-zero} probability in a finite number of iterations), and\\\vspace{-16pt}
    \item the chain is aperiodic. 
\end{enumerate}

For MCMCP experiments, we argue that both assumptions are fairly realistic.
If participants have a \mbox{non-zero} probability of doing something completely unexpected, assumption \textcolor{Blue}{\textit{1}} is satisfied.  
Assumption \textcolor{Blue}{\textit{2}} only requires that a participant be willing to be ``lazy'' every once in a while.  

\vspace{12pt}
\subsection{Rate of Convergence to the Prior}
\label{app:convergence}

How many iterations does it take for a given MCMCP chain to be sufficiently close to the prior? 
For a small number of nodes $n$, it is possible to enumerate all nonisomorphic graphs $\in \GraphSet_n^{ }$.  
In such cases, we can answer the question of how fast the MCMCP chain converges to the prior in terms of the (asymptotic) mixing time, 
which we define as the time it takes for the distribution to get a factor of \mbox{$e\approx2.718$} closer (in total variation distance) to the prior (in the limit of a large number of iterations). 

For a given choice of prior over these graphs, one can explicitly construct a transition matrix $\TransitionMatrix$ representing the composed mapping \mbox{\M{\ParticipantMapping(\ExpMapping(\cdot))\!:P(\GraphSet_{n}^{ })\!\rightarrow\!P(\GraphSet_{n}^{ })}}. 
Then the mixing time \mbox{$\mixingTimeM$} is given by: 
\begin{equation}
    \mixingTimeM = \bigg|\log\!{\Big(\TransitionMatrix(\lambda_2)\Big)}\bigg|^{-1} \label{eq:AsymMixT}
\end{equation}
where $\TransitionMatrix(\lambda_2)$ is the second largest eigenvalue of $\TransitionMatrix$.

\begin{figure}[H]
\vskip 0.3in
	\begin{center}
		\centerline{\includegraphics[trim={0 0 0 1cm},clip,width=0.7\columnwidth]{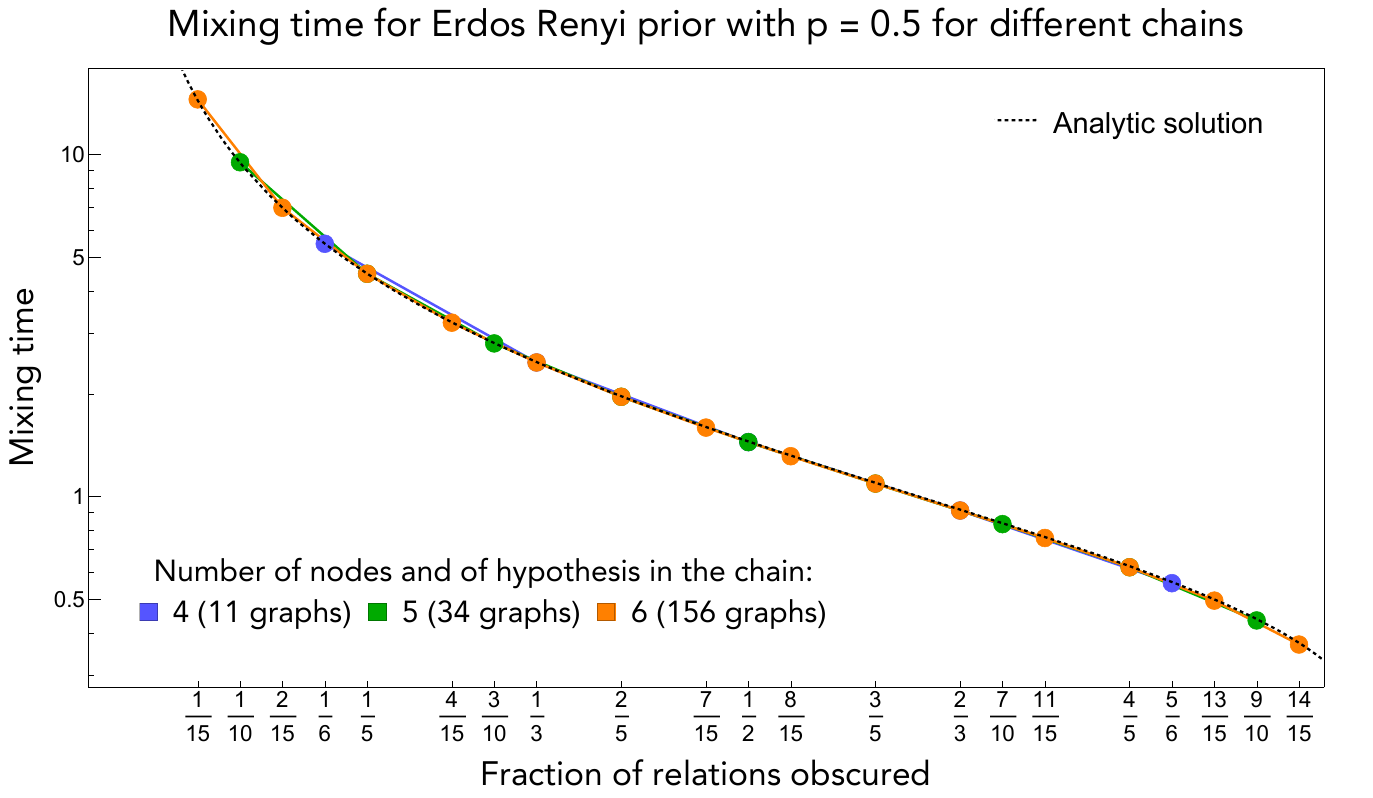}}
	\caption{
 \textbf{Chains converge quickly for \mbox{Erd\H{o}s--R\'enyi}  priors.}\\
Colored markers are the values for the asymptotic mixing times \mbox{$\mixingTimeM$} (eq.~\ref{eq:AsymMixT}), as a function of the fraction of relations obscured $\fractionObscure$, for MCMCP chains over graphs with \mbox{$n=4$} to $6$ nodes and with \mbox{Erd\H{o}s--R\'enyi} ({\ERrhof}) priors (using \mbox{$\rho=\sfrac{1}{2}$})
The dotted black curve is the analytic solution (eq.~\ref{eq:ERAsymMixT}). %
For {\ERrhof} priors, the chains converge to the prior at a rate that depends only on the fraction of relations obscured $\fractionObscure$, 
independent of the number of nodes $n$ and edge density $\rho$.
}
\label{fig:ERuniversalCurve}
\end{center}
\vskip -0.2in	    
\end{figure}

As illustrated in figure~\ref{fig:ERuniversalCurve}, for the simplest case of an \mbox{Erd\H{o}s--R\'enyi} distribution {\ERrhof}, 
the prior converges relatively quickly and convergence depends only on the fraction of relations obscured $\fractionObscure$ 
(independent of the number of nodes $n$ and edge density $\rho$).
The exact expression for the asymptotic mixing time in this case is: 
\begin{equation}
  \mixingTimeER = -\frac{1}{\log\!\big(1-\fractionObscure\big)} \label{eq:ERAsymMixT}
\end{equation}

\begin{figure}[H] 
  \vskip 0.3in
	\begin{center}
		\centerline{\includegraphics[trim={0 0 0 1cm},clip,width=0.85\columnwidth]{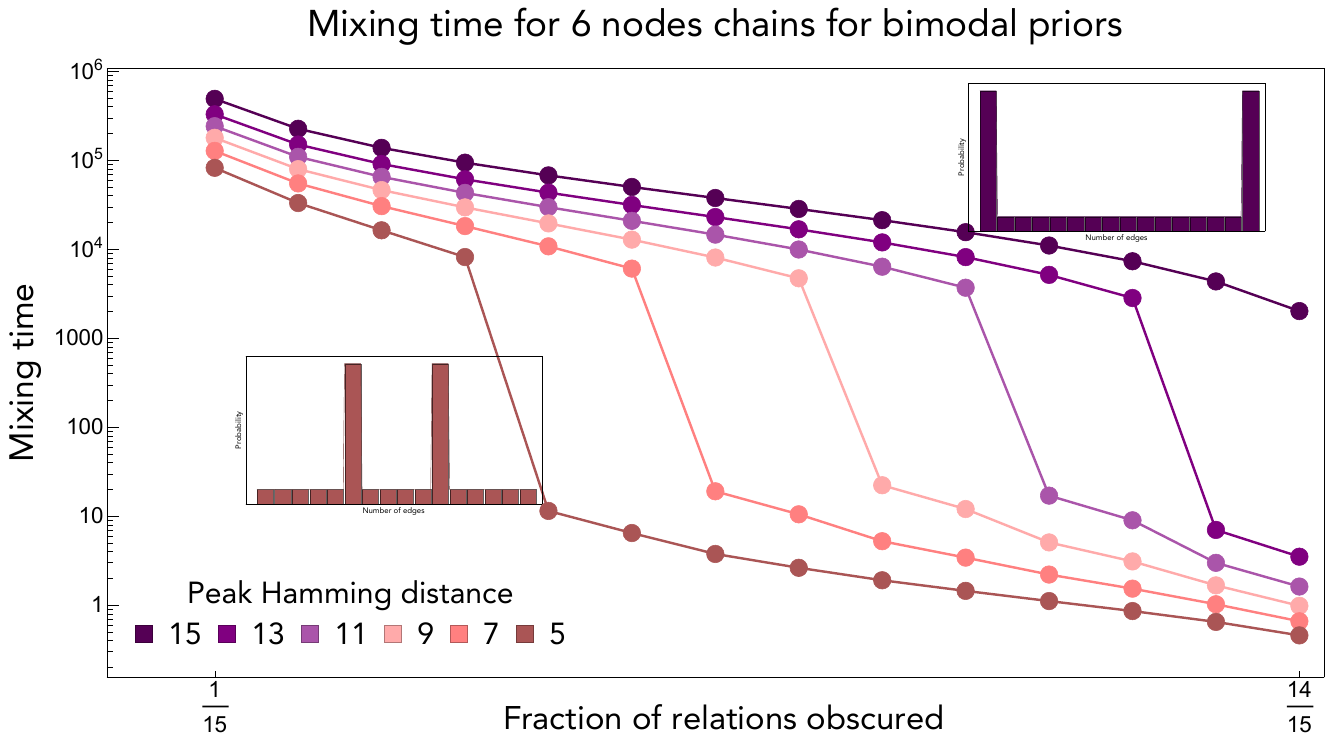}}
 		\caption{\textbf{Convergence rate is highly sensitive to the shape of the prior and the amount of information provided at each iteration.}\\ 
  	We computed the asymptotic mixing times \mbox{$\mixingTimeM$} (eq.~\ref{eq:AsymMixT}) of MCMCP chains over graphs with $6$ nodes for a range of shapes for the priors (different curves) and fraction of relations obscured $\fractionObscure$ (vertical axis).
		We parameterized the priors with probabilities given by the number of edges in the graph (here, $0$ to $15$). 
		In particular, we gave $50\%$ of the probability to graphs separated by some number of edges (the ``peak Hamming distance'') and distributed the rest of the probability uniformly to the other graphs. 
  	E.g., for the prior with a peak Hamming distance of $5$: 
   $25\%$ of the probability is equally distributed between all graphs with $5$ edges, 
   another $25\%$ between all graphs with $10$ edges, 
   and the remaining $50\%$ between all other graphs.  
	As expected, the higher the fraction of relations obscured, the chain mixes faster (small \mbox{$\mixingTimeM$}). 
	Conversely, the larger the distance (in terms of number of edges) between the peaks of the prior, the chain mixes slower. 
 }
	    \label{fig:ConvergenceToPrior}
     \end{center}
\end{figure}

Nevertheless, as illustrated in figure~\ref{fig:ConvergenceToPrior}, for priors with even a moderate amount of structure 
(the priors used in that simulations are only sensitive to the edge distribution),
the mixing time can vary many orders of magnitude depending on the shape of the prior and the number of relations obscured in the partial graph. 

More generally, there is a delicate balance when using MCMCP experiments to recover priors.  
If the experimentalist obscures too little {\InfoInWordP}, they may never know if the experiment sufficiently reached convergence (since the shape of the true prior is unknown). 
On the other hand, obscuring too many {\InfoInWordP} could result in an experiment that is too underconstrained for participants to adequately engage and provide their true prior. 

\newpage
\vspace{12pt}
\section{Advantages of Directly Modeling the Priors in MCMCP Experiments}
\label{app:modelpriorexplict}
In this section, we first demonstrate the advantages of 
explicitly leveraging the assumptions of the MCMCP model 
(section~\ref{app:mcmcpbayesian}). 
We then demonstrate the advantages of our particular choice of model for the priors (section~\ref{sec:priorparameterization}).

\vspace{12pt}
\subsection{Exploiting the Bayesian Assumption}
\label{app:mcmcpbayesian}

As discussed in section~\ref{sec:samplingapproach}, 
the standard approach in MCMCP experiments is to use the observed frequency of the data towards the end of the experiments (once the chain has (hopefully) converged) as a proxy for participants priors.  This wastes much of the collected data. 
In this section, we use some simulations to show that recovering participants priors' by fitting the MCMCP model directly to their aggregated choices indeed uses the experimental data much more efficiently.

To compare these methods, we simulated the responses of ideal Bayesian participants (i.e., respecting all assumptions of the MCMCP model) on our MCMCP experiment over graphs with \mbox{$n=5$} nodes and obscuring half of the relations at each iteration (i.e., \mbox{$\fractionObscure=5$}).
We fit the simulated data using a distribution specifying the probability of all $34$ nonisomorphic simple graphs with $5$ nodes as the model for the prior.  
Our goal with these simulations is to illustrate that, when possible, it is best to recover the prior by directly fitting participants' data to the MCMCP model (regardless of how one might choose to parameterize the prior). 

\begin{figure}[H] 
 \vskip 0.5in
	\begin{center}
      \begin{subfigure}[t]{0.48\columnwidth}
  \centering
    \includegraphics[trim={0cm 0cm 19cm 0cm},clip,width=1\columnwidth]{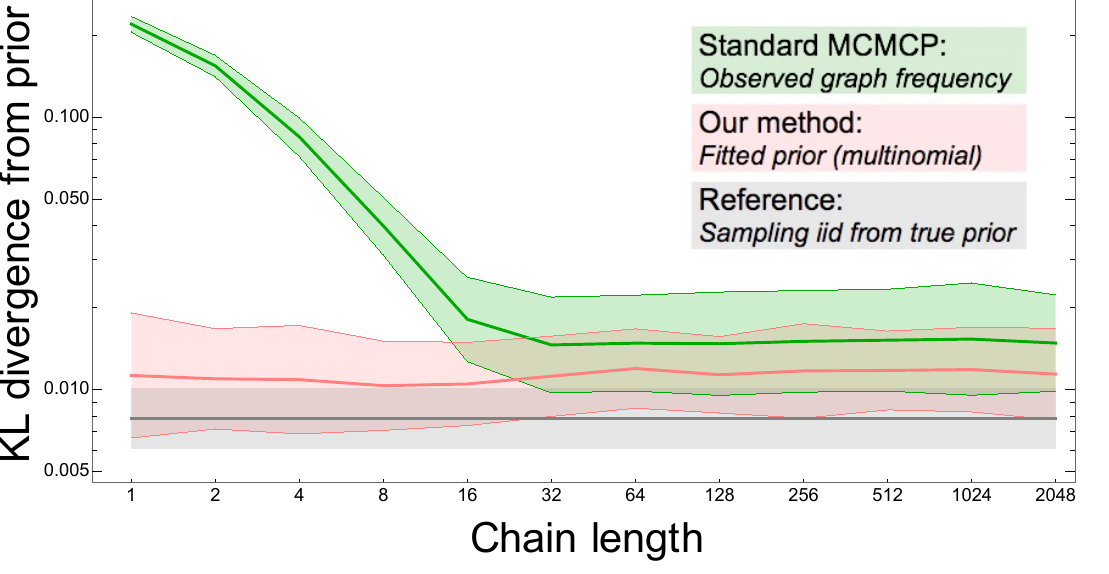}
    \caption{\textbf{Increasing chain length for a fixed amount of data.}\\
    }
    \label{fig:VaryingChainLength}
    \end{subfigure}
     ~
  \hspace{2pt}
    \begin{subfigure}[t]{0.48\columnwidth}
   \includegraphics[trim={19cm 0cm 0cm 0cm},clip,width=1\columnwidth]{Figures/BayesianFittingTwoPicturesMultinomialBeautifulNew.pdf}
   \caption{\textbf{Increasing amount of data for a fixed chain length.}}
     \label{fig:fixedchain}
   \end{subfigure}
  \caption{\textbf{The prior can be more accurately recovered by directly fitting the MCMCP model to the aggregate data.} \\
  We simulated the responses of ideal participants (i.e., respecting all assumptions of the MCMCP model) on our MCMCP experiment over graphs with \mbox{$n=5$} nodes with half of the relations obscured \mbox{$\fractionObscure=5$} at each iteration, 
  using a prior with an asymptotic mixing time of \mbox{$\tau_m\sim14$} iterations. 
	For each simulation, we fit the resulting data by maximum likelihood estimation using a distribution specifying the probability of all $34$ nonisomorphic simple graphs with $5$ nodes as the model for the prior. 
	We then computed the KL divergence from the true prior (used to simulate the data) to: the fitted prior (in \textbf{\color{pink}{pink}});  
	the observed frequency of graphs (in \textbf{\color{green}{green}}); 
	and (as a reference) the distribution obtained by sampling i.i.d.~from the true prior the same number of times (in \textbf{\color{gray}{gray}}). 
	Shading denotes $\pm1$ standard deviation about the mean for $64$ simulations for a given choice of parameters. \\
	\textbf{(a)} For each position in the curve, we varied the length of the simulated chains, but kept the number of data points (i.e., the simulated agents' answers) fixed to $2048$.
	While using the observed frequency is clearly doomed to fail when the length of a chain is shorter than the mixing time $\tau_m$, 
	 fitting the data still does better even when the length is much longer than $\tau_m$. \\
	\textbf{(b)} We kept the chain length fixed to $16$ and varied the number of data points. 
	As the number of data points increases, fitting the prior continues to improve, while using the observed graph frequency asymptotes to some finite error. 
	This asymptote is mainly due to the contribution from graphs in the beginning of the chains. 
 }
    \label{fig:FigFitBoth}
	 \end{center}	
\vskip -0.2in	
\end{figure}

As shown in figure~\ref{fig:FigFitBoth},\footnote{We use KL divergence as a measure of closeness to the prior (as opposed to, e.g., total variation distance), as it is more sensitive to relative differences in probabilities. 
However, the results are similar when using other measures.} fitting the data to the MCMCP model (pink curves) recovers the prior more accurately than the more standard approach of using some of the observed data as a proxy for the prior (green curves).  
This is especially true when the chain lengths are limited (fig.~\ref{fig:fixedchain}).
In addition, the fitting approach does not require estimation of the mixing time, which can vary dramatically depending on the prior, number of nodes, and fraction of relations obscured (see figure~\ref{fig:ConvergenceToPrior}). 

\begin{figure}[H] 
 \vskip 0.2in
	\begin{center}
		\includegraphics[width=0.55\textwidth]{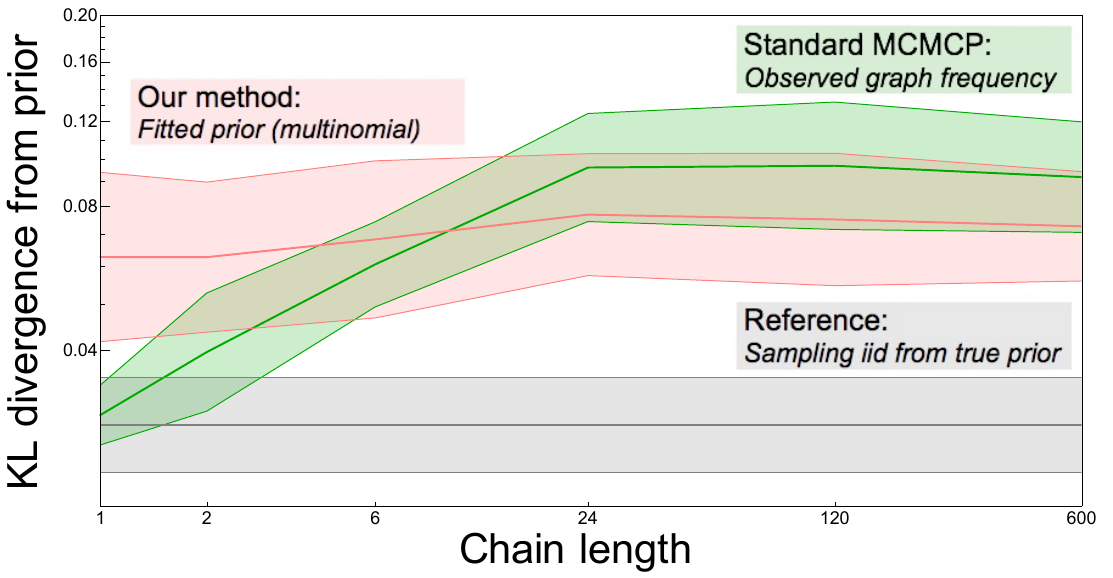}	
\caption{\textbf{The fitting method outperforms the  MCMCP sampling approach, even when the \mbox{``burn-in''} period is eliminated.}\\ 
We generated synthetic data using the same specification as in figure~\ref{fig:FigFitBoth}, 
but starting with artificially \mbox{pre-converged} chains, by initializing the chains using a graph sampled from the true prior.
For each position in the curve we varied the length of the simulated chains, but kept the number of data points fixed to $600$. 
When the chain length is one, the perfect initialization renders the standard approach almost equivalent to sampling i.i.d.~from the true prior. 
However, as the chain length increases, correlations between neighboring samples result in a decrease in the effective sample size, and the error when using the standard MCMCP sampling approach increases.  
When the chain length is \mbox{$\gtrsim\tau_m$}, recovering the prior by directly fitting the MCMCP model to the data outperforms the standard approach.}
	    \label{fig:ChainLengthFixed}
     \end{center}
      \vskip -0.2in
\end{figure}

In figure~\ref{fig:ChainLengthFixed}, we remove issues that are due to the \mbox{burn-in} period by initializing the simulated MCMCP experiments/chains with a graph sampled from the underlying prior of the simulated agents. 
We find that the fitting method \textit{still} outperforms the standard approach, even when the simulated chains have length much longer than the mixing time. Indeed,
even with the \mbox{burn-in} period removed, neighboring data points in a MCMCP chain are correlated.\footnote{Asymptotically, one can approximate the effective sample size by dividing the total number of data points by the mixing time, although more precise estimates exist (for example, see \citet{hsu2015mixing}).}  
This results in a decrease in the effective number of samples obtained from such an experiment. 
 %
\newpage


\newpage

\vspace{12pt}
\subsection{Prior Parameterization}
\label{sec:priorparameterization}

\begin{figure}[H] 
  \vskip 0.2in
	\begin{center}
		 \includegraphics[width=0.55\textwidth]{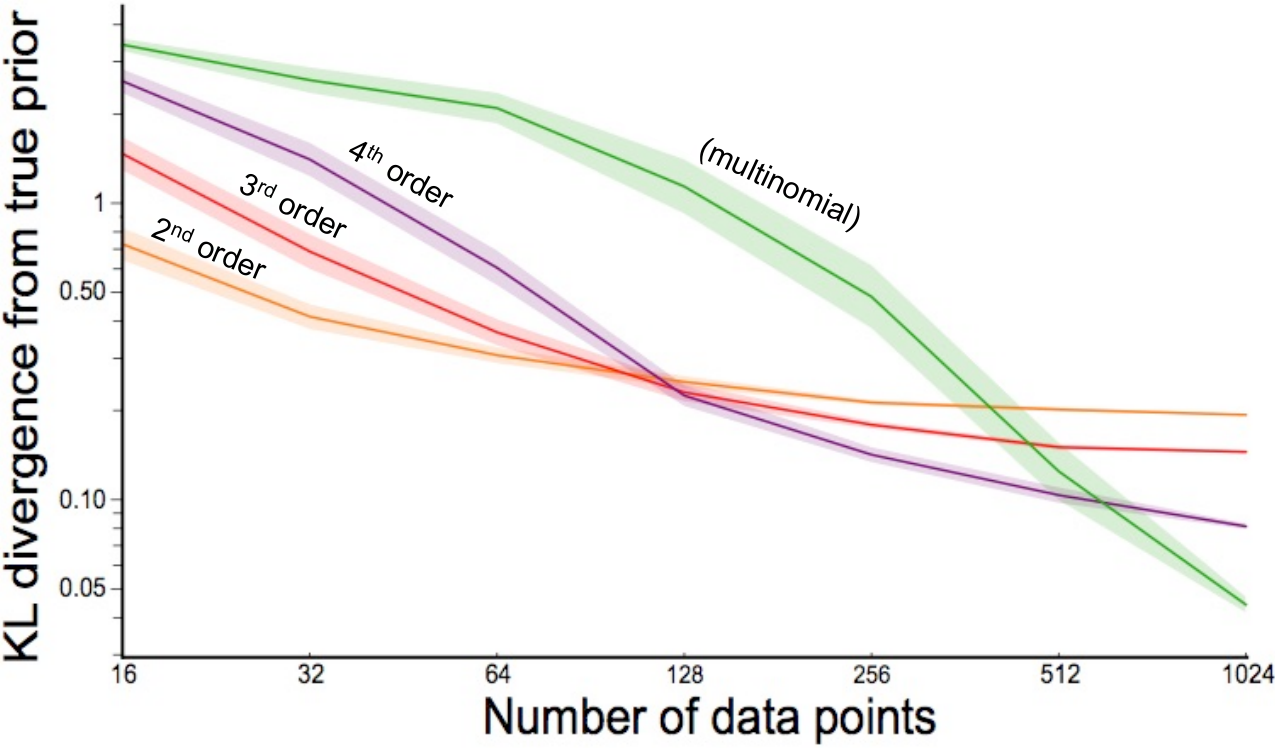}
		\caption{\textbf{Our hierarchical parametrization of distribution over graphs allows for more accurate recovery of the prior.}\\ 
		We simulated data on our MCMCP experiment over graphs with $4$ nodes (there are $11$ nonismorphic graphs in total). 
  	We then fit the MCMCP model to these simulated data using our hierarchical parameterization of the prior (eq.~\ref{eq:LowDimensionalPrior}) for several choices of order $r$.
   Larger $r$ corresponds to more constrained subgraph densities, thus more structured/complex priors.  (For $4$ nodes, \mbox{$r=6$} constrains all subgraph densities). 
	Shading corresponds to $\pm1$ standard error about the mean for $64$ runs of this simulation. 
	When the data are limited, the model with fewer parameters recovers the prior more accurately. 
	As the quantity of data increases, the ordering incrementally inverts until the model with highest complexity does best.  
	However, as the number of parameters in a full multinomial model is \mbox{super-exponential} in the number of nodes (see table~\ref{table:numgraphs}), and engaged human attention is expensive and difficult to obtain, the optimal order will typically be intermediate.}
	 \label{fig:RecoverSyntheticData}
  	\end{center}
\end{figure}

\begin{figure}[H] 
\vskip 0.2in
	\begin{center}
	\centerline{\includegraphics[width=0.55\textwidth]{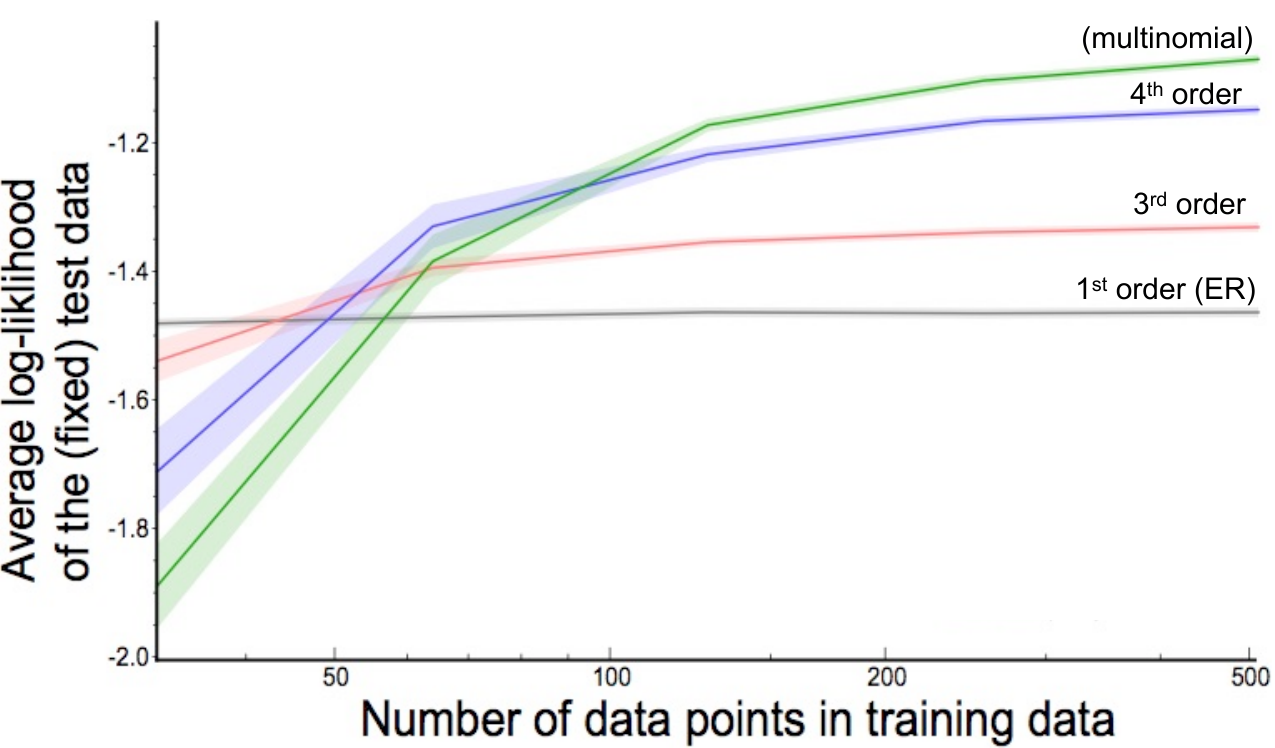}}
		\caption{\textbf{Our hierarchical parameterization of the prior improves generalization in real data.} \\
	We used $1210$ data points from participants doing our experiment over graphs with $4$ nodes. 
 We randomly partitioned the data into test ($698$ data points) and training data. 
 We then fit the MCMCP model to the training data using our hierarchical parameterization of the prior (eq.~\ref{eq:LowDimensionalPrior}) for several choices of order $r$, and evaluated their \mbox{log-likelihood} in the same fixed test data.
	Shading corresponds to $\pm1$ standard error about the mean for $64$ repetitions of this process. 
	In accord with the \mbox{bias-variance} tradeoff, when the data are limited, using a \mbox{lower-order/simpler} model for the prior results in better generalization (i.e., higher \mbox{log-likelihood} of the unseen data). 
	However, as the number of data points increases, \mbox{higher-order/more structured} priors do increasingly better. 
	Again, in practice, the optimal order is typically intermediate.} 
	    \label{fig:GeneralizationInData}
     	\end{center}
      \vskip -0.2in
\end{figure}

In this section, we demonstrate two practical advantages of our hierarchical parameterization for the prior (equation~\ref{eq:LowDimensionalPrior}).
In particular, as shown in figure~\ref{fig:RecoverSyntheticData}, this parameterization results in 
more accurate recovery of the prior in simulated data (where we know the ground truth),
and as shown in figure~\ref{fig:GeneralizationInData}, it also improves generalization in the real data.

\end{document}